\documentclass[b5paper,12pt]{report}    	%%  This sets the printing on both sides of a page
\usepackage{setspace}
\usepackage{geometry}
\usepackage{phdThesisFormat}      							%%  This is the sty file that is to be used to generate the 
\usepackage{times}                							%%  This is the font that the document is to use. This is a must!!!
\usepackage[english]{babel}

\usepackage{amsmath}
\usepackage{graphics}
\usepackage[pdftex]{graphicx}    							%%  This package is necessary to use pdflatex excellent results
\usepackage{url}
\usepackage{amsfonts}
\usepackage{caption}
\usepackage{array}
\usepackage[colorlinks=true, linkcolor=black, citecolor=blue, urlcolor=blue]{hyperref} 
\usepackage{newunicodechar}
\newunicodechar{ﬁ}{fi}
\newunicodechar{ﬀ}{ff}
\advisor{Prof. Dr. Yonghwan Oh}                             			%%  Your advisor's name
\advisorInstitution{Korea Institute of Science and Technology}           			%%  Your advisor's institution
\committeeMemberA{Prof. Dr. Doik Kim}                   	%%  Committee member A's name
\committeeMemberAInstitution{Korea Institute of Science and Technology}    %%  Committee member A's institution
\committeeMemberB{Prof. Dr. Keehon Kim}                   	%%  Committee member B's name
\committeeMemberBInstitution{Korea Institute of Science and Technology} 	%%  Committee member B's institution
\graduateProgramDirector{Prof. Dr. Sang Chul Ahn}        			%%  Graduate program director's name
\researchCenter{Department Chair of HCI \& Robotics}                   								%%  The research center you are associated with
\doctoralStudent{Minh Nhat Vu}         		%%  Very important: Your name
\dissertationMonth{July}                                       %%  The month when you give your thesis
\dissertationYear{2017}                                         	%%  The year when you give your thesis
\dissertationTitle{Hierarchical Control Strategy for Planar Bipedal Walking Robot based on the Reduced-Order model}								%%  The title of your dissertation
\campus{KIST, Seoul}                                            		%%  The campus where you are to give your thesis
\major{HCI \& Robotics}                                      	%%  Your major
\abstractFile{abstractFile}                                     	%%  This is a plain tex file that holds your abstract
\acknowledgementFile{acknowledgementFile}                       	%%  This is a plain tex file that holds your acknowledgements
\dedicationFile{dedicationFile}                                 	%%  This is a plain tex file that holds your dedication
\bibliographyFile{References}                                 		%%  This is a bib file with the bibliography for your dissertation

\usepackage[utf8]{inputenc}

\begin{document}

% The following commands are not to be changed as they prepare the initial pages of the document. Also, they set up the environment for the real content.  Do not move them or erase them; they are needed for a proper presentation of your dissertation

\MainTitle
\ThesisCommittee
%\Declaration
\Dedication
\Acknowledgements
\Abstract
\contentsAndLists
\startContent

% Here you call the files that you prepared in order to get a nice and concise master thesis (i.e., the chapters).  Just call on the command as many times as necessary. Note that it is the same for either thesis chapters or appendices.

\chapter{Introduction}
\label{chapter:chapter01}
\section{Motivation}
In recent years, there have been great advances in robotics, e.g., smart household assistant robots, humanoid robots, and exoskeletons for assisting in rehabilitation. 
\begin{figure}[h]
    \centering
    \includegraphics[width= 0.8\textwidth,height=1\textheight,keepaspectratio]{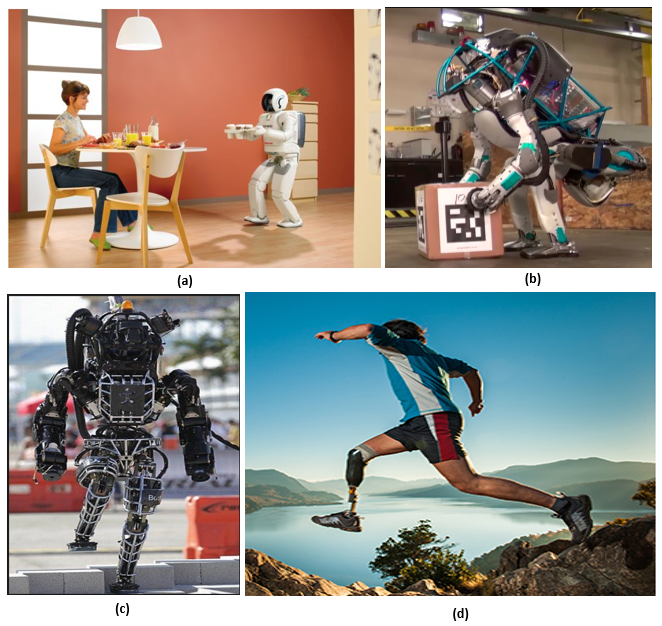}
    \caption{The robot Asimo serves a delicious breakfast (a) . The next generation of Atlas lifts the box (b). The first generation of Atlas can walk on rough terrain (c). The prosthetic limb can recover and improve human strength.}
    \label{figure1}
\end{figure} 
Some humanoid robots are illustrated in Fig. \ref{figure1}. 
Honda has developed Asimo based on its similarity to humans as decipted in Fig. \ref{figure1}a. 
Asimo has the potential to perform human tasks in environments designed for human activities. 
Boston Dynamics has developed the Atlas robot that can walk on a terrain (Fig. \ref{figure1}c) and lift an object (Fig. \ref{figure1}b). 
The prosthetic limb shown in Fig. \ref{figure1}d helps the user regain balance by functioning similarly to human legs. 
These examples have shown the tremendous benefits that robots can bring to humanity
%, and I believe that humanity will maintain an efficient method of controlling robots when they eventually become part of our daily lives. 

%Considering this, my greatest desire is to build reliable, effective, and safe robots that can be easily integrated into our daily lives. 
%In this thesis, 
%as a first step towards achieving my ultimate goal, I would mainly like to develop a controller that enables stable bipedal walking motion in the sagittal plane. 
The controller is an important part of every robotic system. 
Specifically for humanoids, the balance controller plays an important role to maintain the upright posture of the robot. 
In order to develop such a controller, a reduced-order model of the corresponding complex model of the robot is often utilized. 
%and then translate these insights into a target behavior for our controller. 
For this reason, in this thesis, the Biped Trunk Spring Loaded Inverted Pendulum Model (BT-SLIP) is considered to be a reduced-order model of the articulated body robot. 
%and a controller designed for the articulated robots (the 5-link robot and the 7-link robot) will be conducted. 
This reduced-order model, also known as the template model, is designed to match the forces generated by humans during walking. 
This template model can give meaningful information to the control system to reproduce the observed human dynamics, e.g., the ground reaction force model. 
%On the other hand, we thought that in bipedal walking, the trajectories in space need not be precise. 
%Our control system is based on the current state of the robot, and I would also like to answer the question of whether it is possible to develop a simple and reliable method for bipedal walking. 
%This research started with the above question. 
%If a method of bipedal walking was so simple and reliable that it could be easily implemented by many researchers, it would be a very positive contribution to the robotics society. 
%For this reason, in this work, a simple walking method for humanoid robots is presented. 
\section{Literature review}
Controlling a bipedal robot is challenging due to the high number of degrees of freedom (DoFs) and the hybrid nature of stepping, where the continuous model changes at each stage, and can quickly become under-steered. 
Since it is difficult to directly control all of the dynamics, hierarchical control approaches are becoming more popular. 
Simple template models have been shown to be useful for capturing and analyzing animal and human locomotion behavior, and a mapping controller translates the behavior into individual actuator inputs from the articulated robot. 
\subsection{Reduced-Order Models for Biped Robots}  
\subsubsection{The Inverted Pendulum model}
\begin{figure}[ht]
\centering   \includegraphics[width=0.2\textwidth,height=0.5\textheight,keepaspectratio]{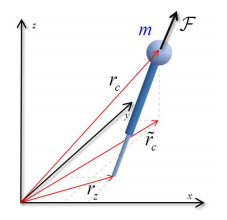}
\caption{ The 3D model of the LIP}
\label{fig:LIPM}
\end{figure}

The inverted Pendulum (IP) \cite{hemami1977inverted} is one of the earliest template models that has its center of mass vaults over a pivot point, constructed by a fixed mass-less leg. 
The IP model is widely applied to analyze the Passive Dynamic Walker \cite{mcgeer1990passive}, \cite{collins2001three}, and human motion \cite{kuo2005energetic}. 
Later, various simple walking robots are presented. Katoh et al. \cite{katoh1984control} built a walking robot whose controller is based on a stable limit cycle of the double IP model. 
Hurmuzlu et al. \cite{hurmuzlu1986role} stated that the ground reaction impacts were a major contributor to dynamic walking stability. 
Later, researchers extended the simplified IP model to the Linear Inverted Pendulum (LIP) model. 
%The LIP model can be used as a model for the analysis of a humanoid robot in 3D space. 
A 2D version of the LIP model was first proposed in 1991 \cite{kajita1991study} and extended to a 3D version \cite{kajita20013d}, as shown in Fig. \ref{fig:LIPM}. 
It consists of assumptions based on approximating the dynamics of the robot by an inverted pendulum in the following:
\begin{itemize}
    \item The robot is represented by a point mass m located at its center of mass (CoM).
    \item The legs of the robot are massless and freely moved in swing phases (it has no swing actuation).
    \item The height of the CoM is kept constant during the motion ($z_c = constant$). 
\end{itemize}

By constraining the CoM movement of an inverted pendulum model in a horizontal plane, the relationship between the position of the Center of Pressure (CoP) and the CoM state can be simplified
\begin{equation}
\begin{aligned}
& p_y = y_c - \dfrac{z_c}{g}\ddot{y}_c \\
& p_x = x_c - \dfrac{x_c}{g}\ddot{x}_c,
\end{aligned}
\end{equation}
where $[p_x,p_y]$ is the position of the CoP and $[x_c, y_c, z_c]^T$ is the position of the CoM. 
The LIP model is widely used by many interesting and successful humanoid dynamic walking controllers. 
Kajita et al. \cite{kajita2003biped} proposed the ZMP Preview Control, which is inspired by Model Predictive Control (MPC). 
They used Model predictive control (MPC) to generate a control based on the predicted future states using a prediction horizon. 
This method takes into account the pre-planned future ZMP reference locations and uses the jerk of the CoM as a control signal. 
Then, a dynamically stable CoM trajectory can be generated by solving an optimal control problem that minimizes the ZMP tracking error and control error while maximizing the CoM trajectory smoothness. 
Later, there are several improvements that are applied to ZMP preview control. 
To deal with uneven ground and external disturbances, Kajita et al. \cite{kajita2006biped} introduced Auxiliary ZMP Control that temporarily forces the ZMP to deviate from the reference trajectory by a certain amount, but there are some negative effects to the subsequent walking motions in an undesirable way.
\subsubsection{The Spring Loaded Inverted Pendulum (SLIP) model}
More recently, the SLIP model is the other well-known model which predicts and explains essential characteristics of human walking and running such as the gait-specific pattern of ground reaction force (GRF) and the center of mass (CoM) trajectory \cite{blickhan1989spring}, \cite{mcmahon1990mechanics}. 
In this model, a whole body weight concentrates in its pelvis position, and its leg acts as a massless spring. 
Inspired by the SLIP model,  Geyer et al. \cite{geyer2006compliant} proposed a walking template which is called the Bipedal Spring Loaded Inverted Pendulum (BSLIP) model. It has some advantages when compares with the LIP model, e.g.,  this model reproduces the CoM vertical oscillations and the double peak pattern of the ground reaction force, which are the essential characteristics in human walking. 
Moreover, the double support phases are included naturally without additional mechanisms required. 
That advantage creates the resemblance from a human-like gait to the walking motion of the BSLIP model and allows it to close of the gap between humans and humanoid robots. Nevertheless, the point-mass assumption hinders this model to address postural control whereas vertical body alignment plays a significant role in the stabilization of human locomotion \cite{maus2010upright}. 
For that reason, the SLIP model is extended to include the upper body (trunk) as in the Trunk-SLIP (TSLIP) model \cite{sharbafi2013robust}, Asymmetric SLIP (ASLIP) model \cite{poulakakis2009spring}, and the Bipedal TSLIP (BTSLIP) model \cite{sharbafi2015fmch}. 
Maus et al. \cite{maus2008stable} proposed the attractive concept for stabilizing the trunk \cite{maus2010upright}. 
The authors introduced Virtual Pendulum (VP) concept through observations in several terrestrial locomotions including humans and then proposed a method to redirect the ground reaction force (GRF) towards a virtual pivot point (VPP) on the trunk located above the CoM. 
The SLIP model with trunk requires the hip-actuated such as the VPP concept and the force modulated compliant hip method (FMCH) \cite{sharbafi2015fmch} to balance the floating body. This model takes into account the ground reaction forces and elastic behaviors similarities with humans and animals. 
\subsection{Task Space Controllers}
Task-Space Control, also known as Operational-Space Control, is a framework that provides flexible and compliant control for highly redundant robotic systems. 
The main focus is on task variables and resolving joint redundancies. 
The tasks can be formulated at velocity, acceleration, or force levels. 
In walking and running locomotion, a task-space controller is often paired with a simple model planner in a hierarchical control scheme \cite{mordatch2010robust}, \cite{wensing2013optimizing}. 
More specifically, at the higher level of control, the simple model planner optimizes the task-space objectives, e.g., the CoM trajectory and foot trajectory, usually with consideration of long-term performance and stability. 
The task-space controller then serves as a low-level control component that focuses on converting the task objectives at the current instant to the whole-body joint space. 

There are different methods to solve task-space control. 
Pratt et al. \cite{pratt2001virtual} proposed the intuitive control framework which is called Virtual Model control. 
In this structure, he used the simple Jacobian transpose to map the forces of virtual springs into joint torques. 
However, this method neglects the Coriolis forces of the system dynamics. 
Later, Park and Khatib \cite{park2006contact} used the nullspace projection methods to solve the task-space control problem. 
A task is defined by equality of reference such as $e = 0$ where $e=s-s^{*}(t)$ is an error to be regulated to 0. 
The task presents a bilateral constraint. On the contrary, the unilateral constraint is typically represented by an inequality $e_i \leq 0$. 
Thus, it is hard to direct this restriction \cite{mansard2008continuous}. 
By planning task dynamics carefully, they could bypass the feasibility issue due to frictional and unidirectional limits on GRF \cite{sentis2010compliant}. 
To avoid calculating null-space projections, task-space control is very often formulated as a multi-objective quadratic programming problem (QP) with constraints on instantaneous dynamics and contact conditions \cite{abe2007multiobjective}, \cite{de2010feature}. 

In recent years, researchers have emerged the centroidal angular momentum as an important task objective for humanoid whole-body motion control \cite{orin2013centroidal}, \cite{orin2008centroidal}. 
It has been shown that properly regulating centroidal angular momentum is crucial in dynamic balancing \cite{hofmann2009exploiting} and highly dynamic movements \cite{wensing2014development}. 
Dai et al. \cite{dai2014whole} optimized the joint trajectories without taking into account the full-body dynamics of the robot. 
The author only considered the simpler centroidal dynamics of the robot. 
This method helps to quickly generate highly-dynamic humanoid whole-body motion plans. 
%%%%%%%%%%%%%%%%%%%%%%%%%%%%%%%%%%%%%%%%%%%%%%% NEW SECTION
\subsection{Hierarchical Control Frameworks}
Bio-inspired templates have been providing insights understanding about locomotion and instructing on possible control strategies for the humanoid robot. 
Based on the viewpoint of energetic and geometric similarities of the human movement to the LIP model, there are several controllers and concepts were proposed. 
Kajita et al..\cite{kajita2006biped} introduced the controller-based zero moment point (ZMP) concept that temporarily forces the ZMP to deviate from the reference trajectory by a certain amount in case of uneven ground or external disturbances during walking. 
Later, "Capture Point" is another concept that is developed with the LIPM model \cite{pratt2001virtual}. 
The capture point is a point on the ground where the robot can step to bring itself to a balanced state. 
Later the capture point concept is extended to as the "Divergent Component Point" (DCM) in \cite{takenaka2009real}. 
Englsberger et al. \cite{englsberger2011bipedal} proposed two control strategies for DCM tracking in the DLR-Biped robot. 
More recently, Englsberger et al. \cite{englsberger2013three} extended the DCM concept to a 3D model and allowed the LIPM to walk reliably over uneven terrain. 
They found that the walking DCM control framework is more flexible than the ZMP preview control. 
So far, the DCM-based walking controller has been applied on many experimental humanoid platforms and proven effective such as Atlas robot \cite{englsberger2015three}, Thor robot \cite{hopkins2015compliant}, and M2V2 robot \cite{pratt2012capturability}. 

On the other hand, ground reaction forces and elastic behaviors are other favorite aspects addressed by another controller based on the SLIP model. 
The SLIP model has been widely used as a motion template for running and hopping (jumping) in biped and humanoid robots. 
For instance, Hutter et. al. \cite{hutter2010slip} proposed the hybrid controllers for the running robot StarETH. 
This controller is proposed to combine the motion of CoM predicted by the SLIP model and Operational Space Control as a higher lever controller. 
Mordatch et al. \cite{mordatch2010robust} approximated the SLIP model by decoupling and linearizing the inverted pendulum in the horizontal plane and adding a spring with constant stiffness in the vertical direction. 
The approximate model has closed-form dynamics, so they can run a population-based preview optimization (based on Covariance Matrix Optimization \cite{hansen2006cma}) in real time to select CoM and foot trajectories for both running and walking gaits. 
Garofalo et al. \cite{garofalo2012walking} embedded the Biped-SLIP dynamics into a five-link biped robot model, but they assumed the feet have full actuation on the ground. In terms of adjusting the stiffness in the SLIP model, Visser et al. \cite{visser2012robust} presented the feedback linearization law so that it will reliably track a precomputed trajectories states of the SLIP model in the presence of a disturbance. 
Hereid et al. \cite{} used the CoM trajectory generated by a Dual-SLIP model as an optimization cost to tune the parameters of their HZD-based walking controller. 
Recently, Rezazadeh et al. \cite{rezazadeh2015toward} also synthesized a stable walking gait in ATRIAS, a bipedal underactuated robot. 
However, they found that directly commanding an underactuated robot to follow a SLIP-produced trajectory can be problematic in the real world. 
Thus, they attempted to detect the essential stabilizing control laws in the reduced-order model (SLIP) that can also maintain their stabilizing effects on the full-order robot. 
\section{Organization and Contribution} 

This thesis contributes to the development of a hierarchical control strategy for the bipedal walking robot over rough terrain and disturbance forces. 
Inspiring by the Virtual Pendulum concept which describes a mechanical behavior rather than an exact movement trajectory, we develop the Biped Trunk-Spring Loaded Inverted Pendulum (BTSLIP) model as the template model for our control strategy to capture a set of dynamic reference features.
Then, these dynamic reference features are transferred into the articulated robot design via a hierarchical control strategy. 

Chapter \ref{chapter:chapter02} introduces the control strategy on the BTSLIP model. We present two methods to achieve robust walking motion against external disturbances. In the first method, we introduce the combination of the legs' stiffness controller and the discrete linear quadratic regulator (DLQR).  
Then, the feedback linearization controller for leg stiffness tracks both the reference vertical position and velocity of the Center of Mass (CoM) while the DLQR helps to adapt the Virtual Pivot Point by tracking the periodic solution of walking. 
In the second method, to avoid the computational burden and model-based information, we propose the second approach, called the force direction controller, to stabilize the walking motion. 
The algorithms presented in this chapter are published in \cite{vu2017control}, \cite{lee2017control}, \cite{lee2017force}. 
%We claim that the direction of the GRF on the CoM is what important factor to consider, rather than the position of the point where the GRF is redirected to, in order to approach the fascinating VP concept.

Chapter \ref{chapter:chapter03} introduces the articulated robot model with a 5-link used for walking demonstration. 
In this chapter, a state-based swing controller and the force direction controller are implemented for the stance leg. 
Additionally, the task-space controller is utilized to transform the behavior of the simple template model into a more sophisticated robot. %Operation Space Control and the classical Jacobian Transpose are presented in this chapter. 
The implemented controllers are verified with a 5-link biped robot on rough surfaces and external disturbances. This chapter is published in \cite{vu2017walking}

Chapter \ref{chapter:chapter04} extends the 5-link robot to the articulated robot model with the Flat foot. 
By adding the ankle control rules, we use the task space controller implemented in the previous chapter to generate the joint torques of the robot. 
The finite state machine is implemented to give the right decision among control phases. 
The flat foot model is tested in Open Dynamics Engine under a rough surface. Finally, a summary and future work are given in Chapter \ref{chapter:chapter06}. 
This chapter is published in \cite{vu2017finite}

\chapter{Control Strategies for Stabilizing the Biped Trunk Spring Loaded Inverted Pendulum}
\chaptermark{}
\label{chapter:chapter02}
\section{Introduction}
In the field of robotics and biomechanics, the interest in bipedal robotic locomotion has accelerated for many years as a result of higher demands for advanced humanoids to serve in military, exploration, industries and daily life. This advancement is based on the effort to understand human locomotion in biomechanical investigations. Several different approaches were proposed including one well-known approach wherein a simple conceptual template models \cite{full1999templates} were demonstrated to be helpful for targeting and analyzing the behavior in the locomotion of animals and humans. One of the earliest models included an inverted pendulum that has its center of mass vaults over pivot point, constructed by a massless leg (i.e. Inverted Pendulum Model- IP) \cite{hemami1977inverted}. Based on that earlier point of view, Passive Dynamic Walkers (PDWs) were analyzed and demonstrated natural walking behavior \cite{mcgeer1990passive}, \cite{collins2001three}. Researchers soon after applied simplified IP model to Linear Inverted Pendulum and combined it with proper modulation of Zero Moment Point concept \cite{sardain2004forces}. That resulted in the success of many position controlled robots like ASIMO \cite{sakagami2002intelligent} which are more complex and have considerably improved mobility.

More recently, the Spring Loaded Inverted Pendulum (SLIP) model is one of the most attractive templates for predicting and explaining essential characteristics of human walking and running such as the gait-specific pattern of ground reaction force (GRF) and the center of mass (CoM) trajectory \cite{blickhan1989spring}, \cite{geyer2006compliant}. In the SLIP model, a whole body weight concentrates in its pelvis position, and its leg acts as massless spring. However, the point-mass assumption hinders this model to address postural control whereas vertical body alignment plays a significant role in stabilization of human locomotion \cite{maus2010upright}. For that reason, the SLIP model is extended to include upper body (trunk) as in the Trunk-SLIP (TSLIP) model \cite{sharbafi2013robust}, Asymmetric SLIP (ASLIP) model \cite{poulakakis2009spring}, and the Bipedal TSLIP (BTSLIP) model \cite{sharbafi2015fmch}. Maus et al. proposed the attractive concept for stabilizing the trunk \cite{maus2010upright}. The authors introduced Virtual Pendulum (VP) concept by observations in several terrestrial locomotions including human and then proposed a method to redirect the ground reaction force (GRF) towards a virtual pivot point (VPP) on the trunk located above the CoM. 

However, with external disturbance, the VP system slowly converges to t steady motion. 
Sharbafi et al. \cite{sharbafi2013robust} presented the combination between the Discrete Linear Quadratic Regulator (DLQR) and the VPP method in hopping motion. Besides, the muscular-skeletal of the human body can adapt to different gaits and terrains by adjusting the leg stiffness. Visser et al. \cite{visser2012robust} applied this advantage to the SLIP model and showed the robustness against disturbances in walking motion. 

In this chapter, we only consider a planar simple bipedal model with massless springy legs. Firstly, we present the combined control strategy for the BTSLIP model walking under external disturbances. The control strategy is proposed by the combination of controlling leg stiffness and the VPP method coupling with the DLQR. The feedback linearization controller for leg stiffness tracks both reference vertical position and velocity of the Center of Mass (CoM) to the desired trajectory. The desired trajectory of CoM is referred from the periodic trajectory of the BTSLIP model with the VPP method. 

Secondly, without precomputed trajectories, we propose a very simple Force Direction Control for bipedal walking with trunk, inspired by the Virtual Pendulum concept. 
%We first argue that having GRF towards a single ﬁxed point such as VPP or DP is not sufficient for maintaining upright posture, even in a simple model with mass-less legs. 
We first argue that having GRF towards a single fixed point, e.g., VPP or DP, is not sufficient for maintaining upright posture. 
Based on this analysis, we propose a new GRF-redirecting control method for hip torques; the law no longer constrains the direction of GRF towards a single point but to the direction which is always providing a restoring moment to the body. 
A dynamic simulation result demonstrates the effectiveness of the proposed method. The simulation results indicate that the proposed method is a promising method for achieving stable and robust bipedal walking.

The remainder of this chapter is arranged as follows. In Section 2.2, we provide brief descriptions of the BTSLIP model consisting of its configuration, states, and dynamics. Section 2.3 presents the integrated control strategy including the detailed control strategies and simulation results. Finally, Section 2.5 highlights the Force direction controller consisting of the proposed control and simulation results. 
We conclude the paper with final remarks and future works in section V.
\section{The Bipedal Trunk Spring-Loaded Inverted Pendulum }
This section introduces the dynamics of the BTSLIP model. The model consists of the torso and two springy legs. It is assumed that the model is planar and it walks on a flat surface. The model parameters are set to match the characteristics of the human, as given in Table 1. 
\begin{table}[h]
\caption{Model parameters for the BTSLIP model}
\label{table_example}
\begin{center}
\begin{tabular}{| m{5cm} | m{3cm}| m{3cm} |}
\hline    
Parameter & Symbol & Value[unit]  \\
\hline
Torso mass & m & 80[$kg$]  \\

Torso moment of inertia & J & 4.58[\(kgm^2\)]\\

Distance hip to torso & \(r_h\) & 0.1 [\(m\)] \\

Distance torso to VPP & $r_{VPP}$ & 0.1 [$m$] \\

Leg rest length & \(L_0\) & 1 [$m$]  \\

Gravitational acceleration & g &9.81[\(ms^{-2}\)]  \\

Angle of attack & \(\alpha_0\) & \(70.6\)[\(deg\)] \\
\hline
\end{tabular}
\end{center}
\end{table}
\begin{figure}[ht]
    \centering
    \includegraphics[width=0.6\textwidth,height=0.7\textheight,keepaspectratio]{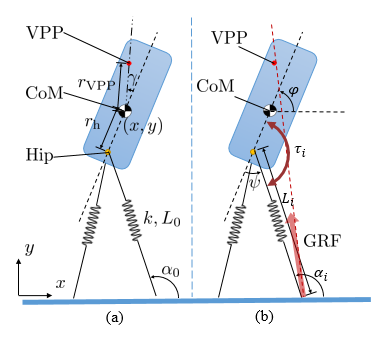}
    \caption{(a) Model parameters. (b) Model variables}
    \label{fig:mesh1}
\end{figure}
\subsection{System Configuration and State Transition}
The BTSLIP model parameters are shown in Fig. 1(a). The model consists of a rigid trunk, which represents the upper body with mass $m$ and moment of inertia $J$, and massless legs. \(L_i\) presents the length of each leg, we indicate the index \(i \in [1,2]\), where 1: left leg and 2: right leg.

The configuration of the system is defined by the variables \(x\), \(y\) and $\varphi$ as the CoM horizontal, vertical positions and the trunk orientation, respectively. These variables, in combination with their corresponding velocities, are used to describe the state of system as \( Q := \{ x_s := [x,y,\varphi,\dot{x},\dot{y},\dot{\varphi}]^T | \: x_s \in o\}\). In particular, $\varphi$ is defined as positive angle in counter clock-wise direction (i.e. $\varphi = 90[\deg]$ if the trunk is vertical). 
 
In a single step, the system can be either in single support (SS) or in double support (DS) which depends on the leg contact conditions. SS begins at the take-off moment of one leg and finishes at the touchdown moment with the same leg. With the assumption of the massless leg, the swing motion of the swing leg does not change the dynamics of the model. The swing leg is assumed to touch the ground with the constant angle of attack, \( \alpha_0 \).

\begin{figure}[ht]
    \centering
    \includegraphics[width=0.35\textwidth]{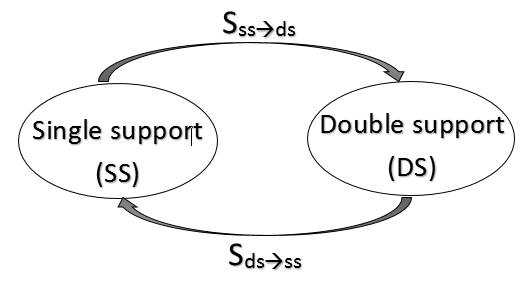}
    \caption{State transition of dynamical system}
    \label{fig:transition}
\end{figure}

The position of hip \( [x_h, y_h]^T\) which is below CoM by \(r_h\), is computed as follows:
\begin{equation}
\label{eqn: hip position}
\begin{aligned}
x_h &= x-r_{h}\cos\varphi \\
y_h &= y-r_{h}\sin\varphi  
\end{aligned}
\end{equation}

The transition from SS to DS can be mathematically defined as: 
\begin{equation}
S_{ss \rightarrow ds} = \{ x_s \in Q \: | \: y_h - L_0 \sin \alpha_0 = 0 \}
\end{equation} 

The position of the foot is defined as  $x_{f,i}$, which is assumed neither slip nor rebound during its stance phase. The length of the stance leg, $L_i$, is calculated by \(L_i = \sqrt{(x_h-x_{fi})^2+y_h^2}\). The transition from DS to SS is written as: 
\begin{equation}
S_{ds \rightarrow ss} = \{ x_s \in Q \: | \: L_i - L_0 = 0 \}.
\end{equation}

During a single step, we define two subsets of \(Q\) accordingly to two switching surfaces above. With \(x_s(t) \: \in \: Q_{ss} \) and  \(x_s(t) \: \in \: Q_{ds}\), the system is in the single support phase and double support phase respectively. Furthermore, a valid walking gait guarantees about not falling (i.e. \( y < 0\)) and forward walking (i.e. \(\dot{x} > 0 \)) in a trajectory \(x_s(t) \: \in \: Q_{ss} \bigcup Q_{ds}\). \(Q_{ss}\) and \(Q_{ds}\) are described below:
\begin{equation}
\begin{aligned}
Q_{ss} = & \:\{ x_s \in Q \: | \: L_0 \sin \alpha_0 < y_h < L_0 \}  \\
Q_{ds} = & \:\{ x_s \in Q \: | \:  0 < y_h < L_0 \sin \alpha_0 \}  
\end{aligned}
\end{equation}

The walking model is called a hybrid dynamical system \cite{sobotka2007hybrid}. Besides the continuous dynamics of the system with a set of ordinary differential dynamics equations, the system has a set of states transition. The transition of the system is shown in Fig. \ref{fig:transition}. 
\subsection{Dynamics of the BTSLIP Model}
\begin{figure}[ht]
    \centering
    \includegraphics[width=0.8\textwidth,height=0.8\textheight,keepaspectratio]{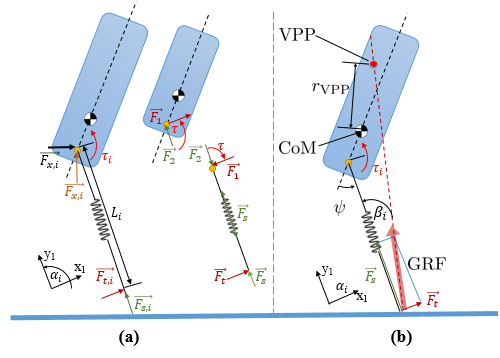}
    \caption{(a) Free body diagram (b) Virtual pendulum-based posture control (VPPC)}
    \label{fig: VPPC}
\end{figure}
In two phases, \(F_{s,i} = k(L_0-L_i)\) gives the spring force along the leg axis where \(L_i, L_0, k\) are respectively the current leg length, leg rest length and the spring stiffness with \(i= 1,2\) for each leg as shown in Fig 1.b. The orientation of the leg $i$ during stance phase, $\alpha_i$, can be trigonometric computed by hip position and the foot contact point.

The dynamic behavior of the system is determined by the forces acting on the CoM. The forces applied to the hip are the sum of the forces generated by spring forces \(\textbf{F}_{s,i}\) and the reaction forces \(\textbf{F}_{t,i}\) which is created by applying the hip torques \(\tau_i\), as shown in (6). A force diagram analysis is depicted in Fig. 3 (a), in local (\(x_1,y_1\)) axis parallel to the leg we have: 

\begin{equation}
\left \{
\begin{aligned}
&\sum \tau = 0, -\tau + F_tL = 0 \\
&\sum F \hat{x}_1 = 0, F_t - F_1 = 0 \\
&\sum F \hat{y}_1 = 0, F_s - F_2 = 0 \\
\end{aligned}
\right .
\end{equation}
Solving (5) for the force applied to the torso (\(F_1\hat{x}_1 + F_2\hat{y}_1\)) as the function of \(\tau\) and \(F_s\), returning the result into the global coordinate yields: 

\begin{equation}
\left \{
\begin{aligned}
F_{x,i} &= F_{s,i} \cos \alpha_i + \dfrac{\tau_i}{L_i} \sin \alpha_i \\
F_{y,i} &= F_{s,i} \sin \alpha_i - \dfrac{\tau_i}{L_i} \cos \alpha_i 
\end{aligned}
\right .
\end{equation}

With \(i \in [1,2]\) as leg index, the dynamical system of the BTSLIP model is described by:
\begin{equation}
\left \{
\begin{aligned}
 m \ddot{x} &= \sum_{i \in \mathbb{C} } F_{x,i} \\
 m \ddot{y} &= \sum_{i \in \mathbb{C}} F_{y,i} -\ m g \\
 J \ddot{\varphi} &= \sum_{i \in \mathbb{C}}^n (\tau_i + r_{COM}(F_{x,i} \sin \varphi - F_{y,i} \cos \varphi))
\end{aligned}
\right.
\label{fig:eomfi}
\end{equation}
where \(\mathbb{C}\) denotes set of contact legs 
\subsection{Searching for Periodic Gait and Local Stability of System}
The search for the periodic solution is taken at the instant of vertical leg orientation (VLO) in SS, and the torso vertically aligns with the foot point. 
Using this definition, the system state at VLO is described with state vector: \(S = (y, \varphi, \dot{x},\dot{y},\dot{\varphi})\). 
The Poincare return map P between two successive VLO is thus defined by \(S_{k+1} = P(S_k)\). 
Based on the Poincare map analysis, we can investigate the orbital stability of a limit cycle. 
For a small perturbation \(\Delta S^*\) around the fixed point, the nonlinear mapping function \(P\) can be transformed regarding Taylor series expansion: 
\begin{equation}
P(S^* + \Delta S^*) \approx P(S^*) + (\nabla P)\Delta S^*
\end{equation}
where \(\nabla P\) is the gradient of P with respect to the states, \(\nabla P\ = \dfrac{\partial P}{\partial S}\bigg|_{S=S^*}\). The local stability of a periodic motion is investigated by computing the eigenvalues \(\lambda\) of the matrix \(\nabla P \). We consider that the periodic motion is stable if all \(\lambda\) are smaller than one \cite{guckenheimer2013nonlinear}. 
\section{The Combined Control Strategy} 
In this section, we propose the control scheme to solve two separated tasks: (a) keeping the upright trunk by the VPP method and DLQR control, (b) tracking the desired vertical position and velocity of CoM. Note that the DLQR control will be active at every VLO event. The stiffness of legs are considered to be external control inputs to the system, and we model the leg stiffness \(k_i\) as the sum of a constant stiffness \(k_0\) and a variable component \(u_i\), i.e. \(k_i = k_0 + u_i \). Hence, the force exerted by the stance leg is given by: \( F_{s,i} = F_{si_0} + F_{si_u}\)
\subsection{The VPP Method and DLQR Controller}
The key idea of the VPPC is to redirect the GRF vectors towards a point located above the CoM, as shown in the Fig.\ref{fig: VPPC} (b).  Hence, the trunk oscillates like a pendulum mounted at the point VPP. By applying a torque \(\tau\) at the hip during walking, that will lead to generate a force perpendicular to the leg axis \(F_N\) to redirect of the GRF vector. As massless leg, the hip torque $\tau_i$ is computed by using geometric relation as:
\begin{equation}
\label{eqn: VPPC}
\left \{
\begin{aligned}
\tau_i &= (F_{si_0} + F_{si_u}) L_i \tan \beta_i \\
\tan \beta_i &= \dfrac{r_{h} \sin \psi + r_{VPP} \sin(\psi-\gamma)}{L_i + r_{h} \cos \psi + r_{VPP} \cos (\psi - \gamma) } 
\end{aligned}
\right .
\end{equation}
where \(r_{h}\) and \(r_{VPP}\) as distance from CoM to hip and VPP, \(\Psi\) and \(\gamma\) are the angle between hip-CoM to leg and CoM-VPP, respectively. \(i \in [1,2]\) is leg index. Solving (\ref{eqn: VPPC}) as the function of \(F_s\) and substituting into (\ref{fig:eomfi}), the equations of motion can be written in more compact form:
\begin{equation}
\dot{x_s} = f(x_s) + \sum_{i\in\mathbb{C}} g_i(x_s) u_i.\\
\end{equation}
where \(\mathbb{C}\) denotes set of contact legs and the input is only \(u_i\).

The DLQR control allows the adaptation of VPP position once per period using the computation which is based on discrete linear quadratic regulator (DLQR). At each VLO, the new VPP position determining by \(r_{VPP}\) and \(\gamma\) is computed for the next phase. Suppose that we have identified a pair \((S^*, \delta^*\)) that will produce a periodic walking gait in the BTSLIP model with the VPPC, i.e. \(r_{VPP}\) and \(\gamma\) are considered as inputs \(\delta = [r_{VPP},\gamma]^T\). We define the change of variables \(\Delta S_n = (S_n - S^*)\) and \(\Delta \delta_n = \delta_n - \delta^*\). The index n denotes the variables in the \(n^{th}\) VLO.

To design the controller, this change of variables allows to analyze the system using the first order of the Poincare return map around the fixed point \(S^*\):
\begin{equation}
\Delta S_{n+1} \approx J_S\Delta S_n + J_\delta \Delta \delta_n 
\end{equation}
where \(J_S =  \dfrac{\partial P}{\partial S}(S^*,\delta^*)\) and \(J_\delta = \dfrac{\partial P}{\partial \delta} (S^*,\delta^*)\). Consider the quadratic cost: 
\begin{equation}
\begin{aligned}
& \min\limits_{\Delta u} \sum_{n=1}^{\infty} \Delta S_n^T Q \Delta S_n + \Delta \delta_n^T R \delta_n  \\
& s.t \:\Delta S_{n+1} = J_S\Delta S_n + J_\delta \Delta \delta_n 
\end{aligned}
\end{equation} 
If the pair (\(J_S, J_\delta\)) is controllable, then the \(\Delta S_k\) can be controlled by using the feedback law as: 
\begin{equation}
\delta_n = -K S_n + \delta^*
\end{equation}
The state feed back gain \(K\) is given by: 
\begin{equation}
K = (J_S^T P J_\delta + R)^{-1} J_\delta^T P J_S, 
\end{equation}
where P is the unique solution of the Discrete- Time Algebraic Riccati Equation (DARE) \cite{bertsekas1995dynamic}.
\subsection{Feedback Linearization Control}
%Inspired by only a certain amount of stiffness variation, adjusted by control effort, it is require to track the desired gait. Thus, once the desired gait is reached, no input force is required to keep the gait [11]. This is enforce that desired gait defines a trajectory in the total state space, specifically in both term of position and velocity. The horizontal position desired gait is identified as a monotonically increasing in time value. Hence, we can parameterize the desired BTSLIP gait by \(x\). Accordingly, desired gait can be depicted as two reference function \(\bar{y}(x)\) and \(\dot{\bar{x}}(x)\). 
%Hence, by using VPPC, which the function of \(\tau\) in each instant of walking is relevant \(u_i\) function, we can rewrite the dynamic by replacing the \(\tau_i\) to the function of \(u_i\) in dynamics system using (8). Dynamics system is equivalent written in more compact form with VPPC: 
%\begin{equation}
%\dot{x_s} = f(x_s) + \sum_{i=1}^2 g_i(x_s) u_i\\
%\end{equation}
%Because of CoM position influences touchdown and takeoff event, the convergence of CoM position becomes more important. In SS phase, system is underactuated which means it only has one control input for converging horizontal position of CoM. In DS phase, the system is fully actuated with two stiffness forces input for converging both position of CoM  and forward velocity. The output target error function is defined as:
From the previous research \cite{maus2010upright} that the system described by (\ref{fig:eomfi}) with a particular set of values for the parameters and the VPPC seems to be suitable for describing human gait and may be used as a basis for more detailed models of human locomotion, referred to as the desired gait. We control the legs stiffness by feedback linearization in order to stabilize the walking model to desired gait. 

Because the CoM position influences touchdown and takeoff event, the convergence of CoM motion to the desired gait trajectory becomes critical. The model CoM horizontal position (\(x\)) of the desired gait trajectory obtained with the VPPC is monotonically increasing with respect to time. Hence, we can parameterize the desired trajectory by $x$. Accordingly, the desired gait trajectory can be described by $\bar{y}(x)$ and $\dot{\bar{x}}(x)$, which are the CoM vertical position and the CoM horizontal velocity. In SS, the system has only one control input which we can utilize to control the vertical position. In DS, we can exploit two control inputs to control the CoM horizontal velocity. Therefore, the output target error function is defined as:

\begin{align}
\begin{bmatrix}
h_1(x_s) \\
h_2(x_s)
\end{bmatrix} 
= \begin{bmatrix}
y - \bar{y}(x) \\
\dot{x} -\dot{\bar{x}}(x)
\end{bmatrix}
\end{align}

\textbf{\(\ast\) For \(x_s \in Q_{ss}\)\(\:\) :} we have the output \(y_s =h_1(x_s)\) results in the second- order input-output dynamics:
\begin{equation}
\begin{aligned}
\dot{y}_s(1) &= \dfrac{\partial h}{\partial x_s} [f(x_s) + g_i(x_s) u_i] =L_f h_1 + L_g h_1 u_i \\
\ddot{y}_s(1) &= L_f^2 h_1 + L_{g_i} L_f h_1 u_i, 
\end{aligned}
\label{eqn:fb1}
\end{equation}
where \(L_g h_1 u_i = 0\), \(L_f^2 h_i, L_f h_i\), and \(L_{g_i} L_f h_i\)  denote the Lie derivatives of the output target function along the vector fields of dynamics system. The feedback linearization control for legs stiffness, 
\begin{equation}
u_i = \dfrac{1}{L_{g_i}L_f h_1} (- L_f^2 h_1 - k_1 L_f h_1 -k_2 h_1)
\end{equation}
% \(u_i\) with \(i= 1, 2\), depends on whether the leg touch the ground. 

By defining the feedback controller in (\ref{eqn:fb1}), the output function are described by second-order form as follows: 
\[
\left \{
\begin{aligned}
& \ddot{h}_1 + k_1 \dot{h}_1 + k_2 h_1 =0 \\
& h_1 = c_1 e^{\lambda_1 t} + c_2 e^{\lambda_2 t} 
\end{aligned}
\right .
\]
If \(k_1, k_2\) are positive constants, then \(Re(\lambda_1)\) and \(Re(\lambda_2)\) are both negative. Therefore, the convergence of \(h_1\) function is guaranteed on \(Q_{ss}\).

\textbf{\(\ast\) For \(x_s \in Q_{ds}\)\(\:\) :} We have fully target function with two control objectives. Because \(L_{gi} h_2 \neq 0\), we can establish the  first-order feedback form for \(h_2\). The feedback control law is defined as below: 
\begin{equation}
\begin{aligned}
\begin{bmatrix}
u_1 \\
u_2 
\end{bmatrix}
= A^{-1} 
\begin{bmatrix}
-L_f^2 h_1 -k_1 L_f h_1 - k_2 h_1 \\
-L_f h_2 - k_3 h_2
\end{bmatrix}
\end{aligned}
\end{equation}
where 
\[
\begin{aligned}
A = 
\begin{bmatrix}
L_{g_1} L_f h_1 & L_{g_2} L_f h_1\\
L_{g_1} h_2 & L_{g_2} h_2
\end{bmatrix}
\end{aligned}
\]
The Lie derivative of \(L_{g_i} L_f h_1\) is not equal to zero as long as the length \(L_i\) of the stance leg $i$ satisfies \(0 < L_i <L_0\). Unfortunately, when the leg touches down the ground, the length of swing leg at this instant is still equal to \(L_0\). That leads $A$ to be singular at a very short finite time. Hence, with this condition, the feedback control with \(L_i = L_0\) in DS phase is defined as: 

\begin{equation}
\begin{aligned}
\begin{bmatrix}
u_1 \\
u_2
\end{bmatrix}
=
A^+ 
\begin{bmatrix}
-L_f^2 h_1 - k_2 L_f h_1 - k_2 h_1
\end{bmatrix}
\end{aligned}
\end{equation}
where \(A^+ = [L_{g_1} L_f h_1 ; L_{g_2} L_f h_1]^+\) is  Moore-Penrose pseudo inverse of $A$.
During DS, using the feedback control law, we can describe the target function \(h_2\) as: 
\[
\left \{
\begin{aligned}
& \dot{h}_2 + k_3 h_2 = 0 \\
& h_2 = e^{-k_3 t} h_2
\end{aligned}
\right .
\]
If \(k_3\) is positive constant then \(h_2\) certainly converges to zero on \(Q_{ss}\). 
\begin{figure}[h!]
    \centering
    \includegraphics[width=0.4\textwidth,height=0.4\textheight,keepaspectratio]{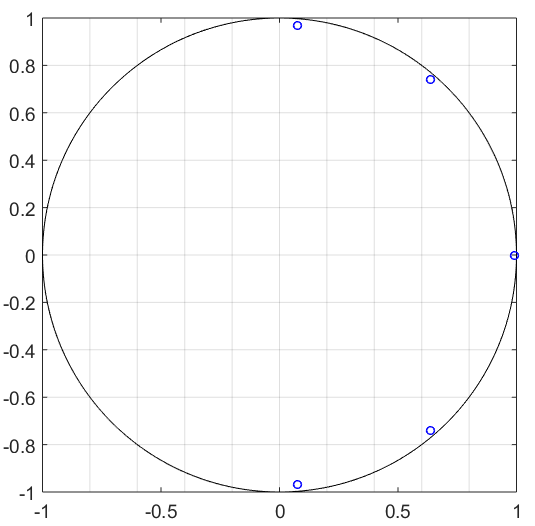}
    \caption{Eigenvalues of the Poincare map linearized around the fixed point of the desired gait.}
    \label{eigen}
\end{figure}
\subsection{Simulation Result of Combined Control Strategy}
In this section, nonlinear dynamic simulation results are presented that validate the control strategy. For the simulations, the BTSLIP model described in Section II is used, together with the control structure as shown in Section III. The main analyzing method for stability of the desired gait is investigating eigenvalues of Poincare return map. Moreover, by adding the disturbance forces are employed to the model to evaluate the robustness.
\subsubsection{Local Orbital Stability}
The numeric simulation shows that the desired gait obtained with a particular set of parameters and the VP model has an asymptotically stable limit cycle in Fig. \ref{eigen}. The eigenvalues are computed as \(\lambda = 0.9992, 0.074 \pm  0.9693i, 0.6387 \pm 0.7400i \). we consider that if the \(\lambda\) have magnitudes less than one, the periodic orbit is asymptotically stable. 
\begin{figure}[h!]
    \centering
    \includegraphics[width=0.85\textwidth,height=0.85\textheight,keepaspectratio]{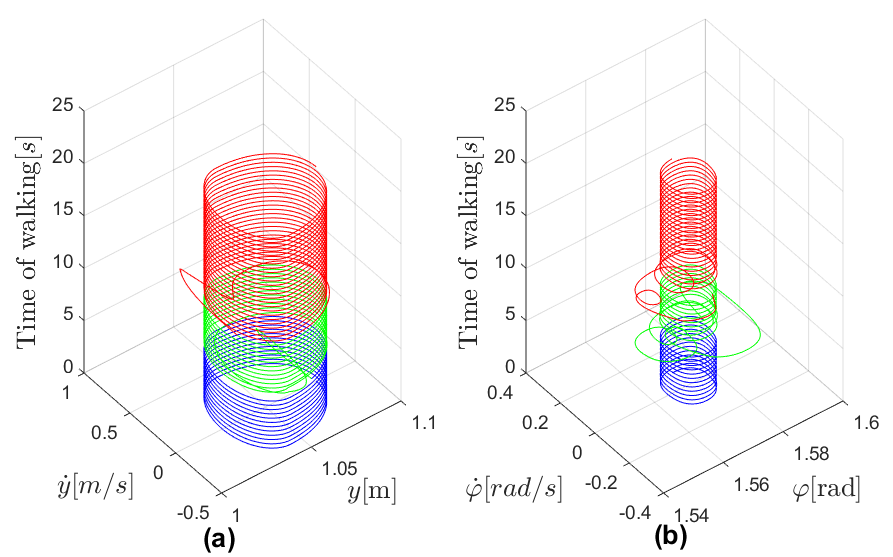}
    \caption{Phase portrait of the BTSLIP model with proposed control under the disturbances. The phases plot are changed color to green and red after the disturbances are applied at \(t = 5s\) and \(t = 10s\). Phase portrait of the vertical position of CoM (a) and phase portrait of the trunk angle (b).}
    \label{proposed}
\end{figure}
\subsubsection{System Responses Against Disturbances}
\begin{figure}[t]
    \centering
    \includegraphics[width=0.85\textwidth,height=0.85\textheight,keepaspectratio]{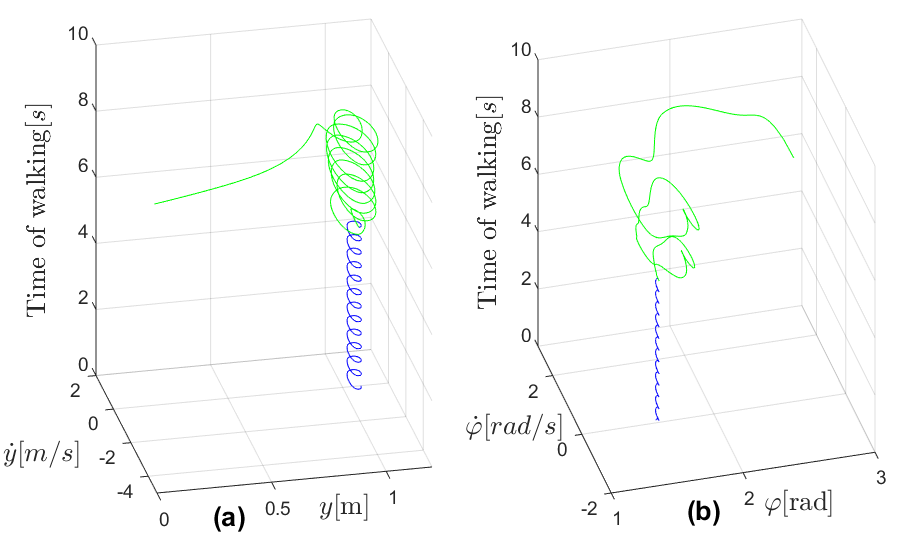}
    \caption{Phase portrait of the VPP method under the disturbances. The phases plot are changed color green after the disturbances are applied at \(t = 5s\). Phase portrait of the vertical position of CoM (a) and phase portrait of the trunk angle (b).}
    \label{VPP}
\end{figure}

In order to demonstrate the effectiveness of the control strategies, we applied the perturbation forces \(F = [F_x,F_y]^T = (-100,300)^TN \) at \(t = 5s\) and \(t = 10s\) in \(0.2\) seconds. Both translation motion and rotation motion were affected by disturbances in the proposed method and the VPP method, as shown in Fig. \ref{proposed} and Fig. \ref{VPP}, respectively. 

Fig. \ref{proposed}(a) presented the phase portrait of the vertical CoM motion. When the disturbance is applied, jerky behaviors are observed before the system response directly returns to the desired periodic orbit by the feedback linearization controller. Fig. \ref{proposed}(b) shows the motion of trunk, which shows much variability after applying disturbance but quickly converges to the steady motion. In comparison with the proposed control strategy, Fig. \ref{VPP} indicates that the VPP method could not reject the disturbance, the BTSLIP model failed after applying the first disturbances at \(t = 5s\).

\begin{figure}[ht!]
    \centering
    \includegraphics[width=0.8\textwidth,height=0.8\textheight,keepaspectratio]{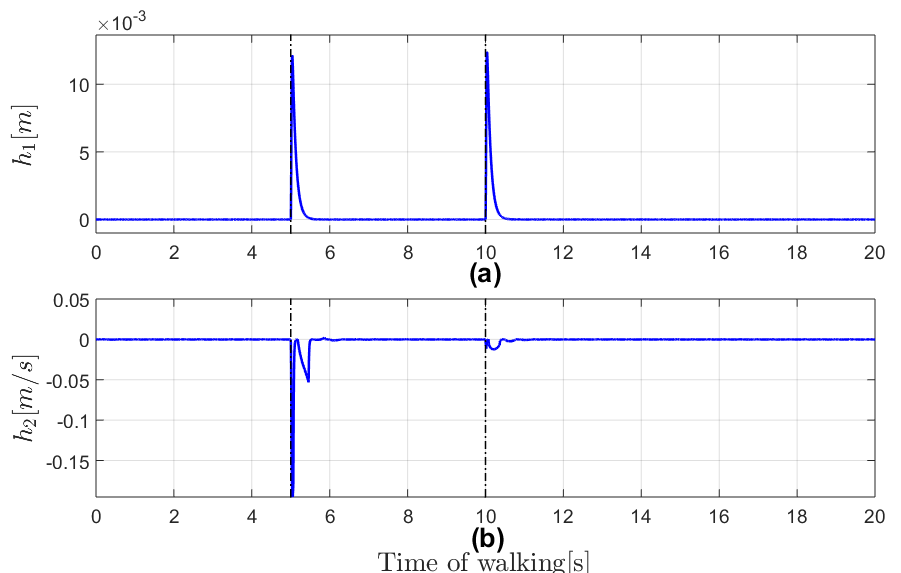}
    \caption{Error output function. The disturbances are applied at the instance of the dashed line. In particular, \(h_1(t)\) exponential decrease to zero in both phase (a), \(h_2\) is directly controlled in DS phase (b)}
\label{error}
\end{figure}

Fig. \ref{error} shows the time progression of the error function \(y_s\). With feedback linearization controller, the error function \(h_1\) exponentially converges to zero in both DS and SS. \(h_2\) performs the exponentially converges to zero in DS.
\begin{figure}[ht!]
    \centering
    \includegraphics[width=0.8\textwidth,height=0.8\textheight,keepaspectratio]{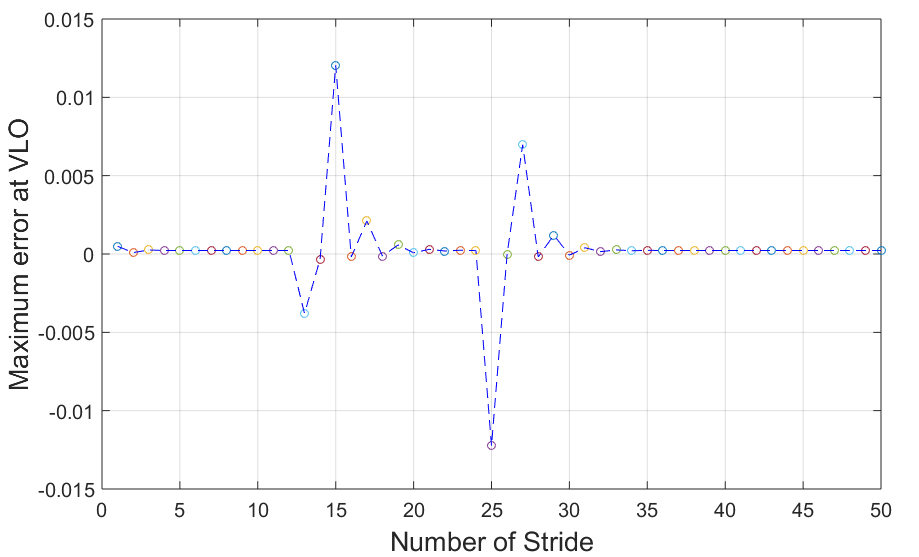}
    \caption{Maximum error at VLO over number strides of walking}
    \label{VLO}
\end{figure}
Fig. \ref{VLO} presents the maximum value of the \(\Delta S\) at \(n^{th}\) stride. It obviously shows that the proposed control can reject the disturbance. We consider that the rejection is satisfied if the gait is sustained around ten steps after the disturbance has been applied.  
\begin{figure}[ht!]
    \centering
    \includegraphics[width=0.8\textwidth,height=0.85\textheight,keepaspectratio]{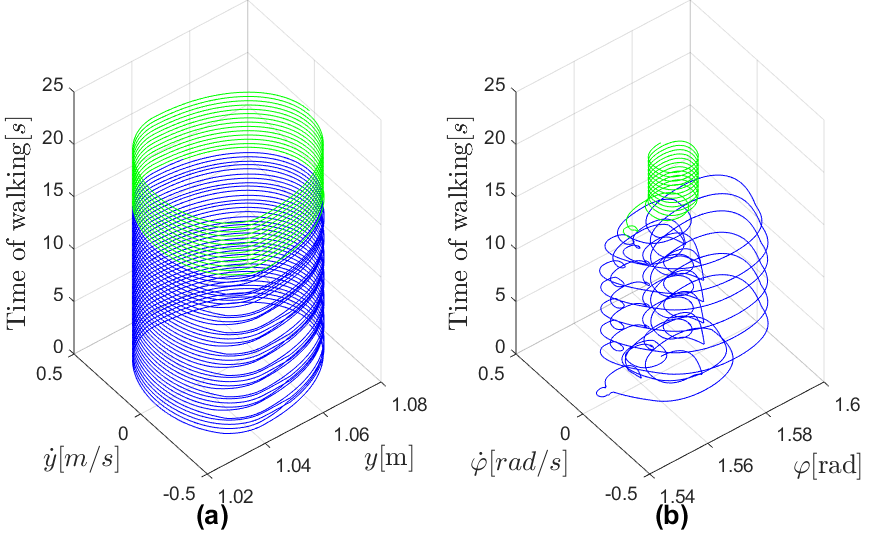}
    \caption{Phase portrait of the BTSLIP model with proposed control under rough terrain (the maximum of height \(h_{max} = 2cm\)). The phases plot are changed color when the rough terrain stop at \(t = 15s\). Phase portrait of the vertical position of CoM (a) and phase portrait of the trunk angle (b).}
    \label{height_change}
\end{figure}

To further validate the controller, the Fig. \ref{height_change} shows performance of the walking model on the rough terrain (height of ground: \(h \leq 2cm\)). The vertical position is immediately stabilized, the trunk motion oscillated with a wide range, but the model still walked. After rough terrain (i.e. $t>15s$), the trunk movement quickly converges to the steady motion. 
\section{The Force Direction Controller}
The previous combined control strategy showed the stability in term of not falling and robustness against the disturbances. However, the controller is depended on the precomputed trajectory from the BTSLIP model with the VP concept. To keep inspiring from the VP concept, we seek a method to control the GRF directions that facilitate stable gait without the need for precise computed trajectory. We first argue that having GRF towards a \textit{single fixed point} (either VPP or DP) is not emerging for maintaining upright posture than adjusting the direction of GRF in an appropriate case; a careful but simple analysis supports this argument. Based on this analysis, we propose a new GRF-redirecting control law for that virtual hip torques should aim; the law no longer constraints the direction of GRF towards a single point but to the direction which is always providing a restoring moment to the body. The method extremely increases the robustness of the system. 

\begin{figure}[tb]
    \centering
%    \framebox{\parbox{3in}{ 
     \includegraphics[width=0.8\textwidth,height=0.85\textheight,keepaspectratio]{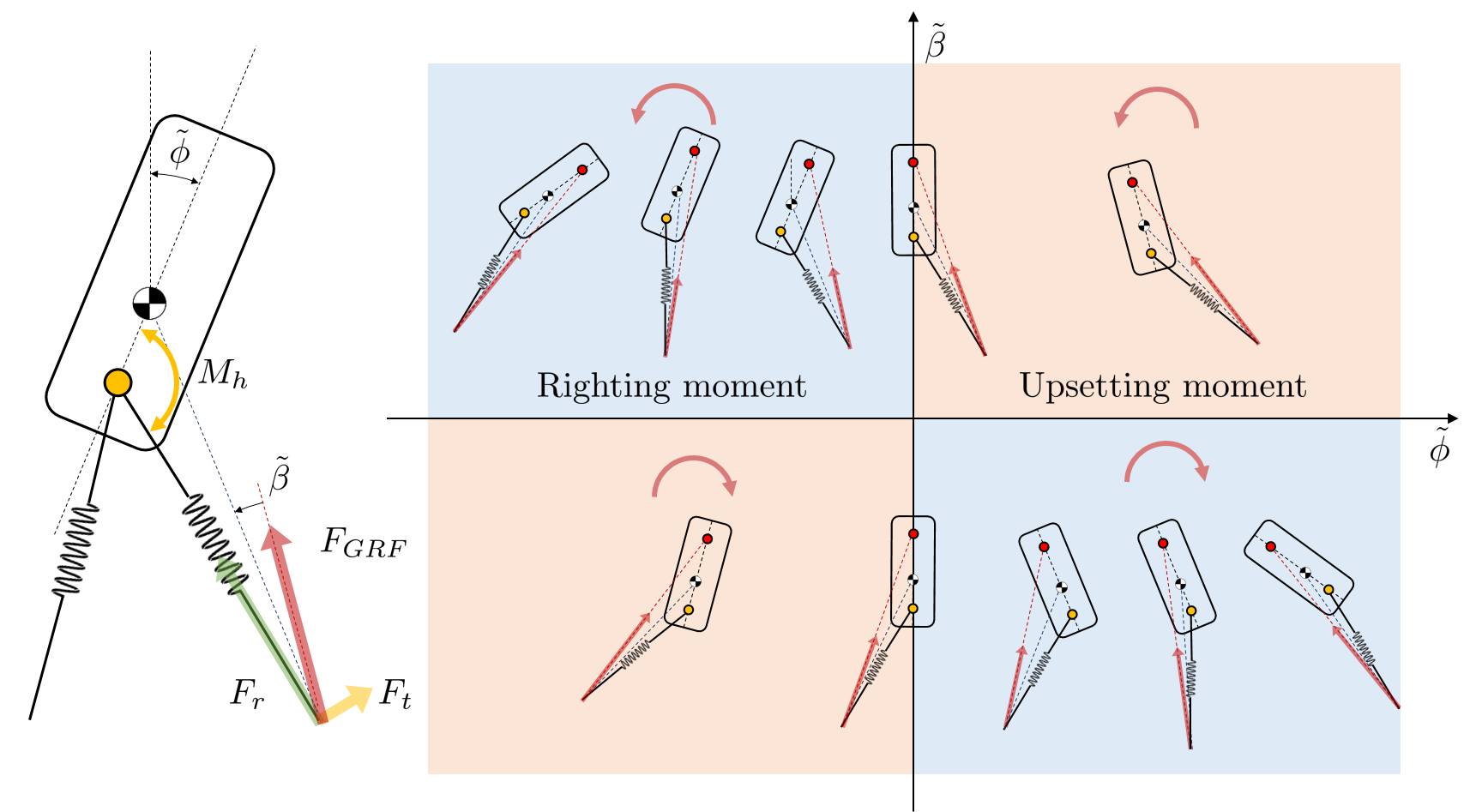}
        \caption{The VPP model with the fixed VPP is not enough for the stable upright trunk. Although usually the GRF will provide restoring moment, in some cases the strategy will provide no restoring moment or even upsetting moment to the body. The graph shows that the region corresponds to $\tilde{\phi}\cdot \tilde{\beta}\geq0$. The variables $\tilde{\phi}$ and $\tilde{\beta}$ are depicted with the model in left.}
    \label{fig2: restoring and uprighting moment}
%    }}
\end{figure}

In Fig. \ref{fig2: restoring and uprighting moment}, we present all possible postures of the schematic model (trunk with massless legs). 
The trunk may be upright or tilted in clockwise or counter clockwise (CW/CCW) direction. At the same time, the foot of stance leg (leg in touch with the ground) may be located right below or posterior/anterior the hip. The combination provides different postures to analyze. Mostly, as intended, the GRF pointing the VPP provides restoring moment, or at worst, zero moments. This allows the trunk be to settle down in some region without specifying the desired posture. However, in some cases, the GRF pointing the VPP provides an upsetting moment, which would cause the trunk to fall. 
In order to provide mathematical criterion, we introduce two variables $\tilde{\phi}$ and $\tilde{\beta}$, pitch orientation from upright posture and the angle between GRF and a virtual line connecting point of action and COM, respectively, as in Fig. \ref{fig2: restoring and uprighting moment}. Then, it can be shown that the cases which collapse the trunk can be represented by the following condition,
\begin{align}
\tilde{\phi}\cdot\tilde{\beta} \geq 0,
\end{align} 
and this can be represented in the graph. 

For example, if we compare a posture in the first quadrant and the right-most posture of the fourth quadrant of the graph in Fig. \ref{fig2: restoring and uprighting moment}, in both postures the trunk is tilted in clock-wise direction and foot is anterior to the hip. However, if the controller tends to generate GRF towards a fixed VPP above the hip, one in the fourth quadrant would provide a moment in a counter-clockwise direction, which is a righting moment, whereas that in the first quadrant in a clockwise direction, which results in an upsetting moment. From this analysis, we end up with that the VPP model is not sufficient for stable and robust upright trunk walking, although the VP concept itself is fascinating, and the important factor is the direction of GRF to be always in the direction of righting moment.  

Experimental findings say that a single intersection point is not necessary for a VP system and cannot always be found \cite{maus2010upright}. Maybe that is because the single intersection point is \textit{insufficient}.

\subsection{Proposed Control Strategy}
\begin{table}[t]
\caption{Control parameters for the simulation}
\label{table:Control parameter}
\begin{center}
\begin{tabular}{|c|c|c|c|c|}
\hline
Parameter & Meaning  & Value [unit]\\
\hline
$r_{VPP}$ & distance between VPP and CoM & 0.1 [$m$]\\
$\alpha_0$ & fixed leg touchdown angle in sagittal plane & 110 [$deg$]\\
$c$ & position-proportional gain & 10[$\cdot$]\\
$d$ & velocity-proportional gain & 1 [$s$] \\
$\mu$ & VBLA parameter & 0.5 [$\cdot$]\\
\hline
\end{tabular}
\end{center}
\end{table}
Although the fixed VPP is not enough, the general argument of the VP concept is still fascinating, in that, the direction of force generated at foot is important in upright trunk walking. We just do not think that a single intersection point is a core of the VP concept. Therefore, we keep the format of the controller such that the hip torque redirect the force generated by the springy leg. The difference is how to define the angle $\beta$. 

\begin{figure}[tb]
    \centering
%    \framebox{\parbox{3in}{ 
    \includegraphics[width=0.8\textwidth,height=0.8\textheight,keepaspectratio]{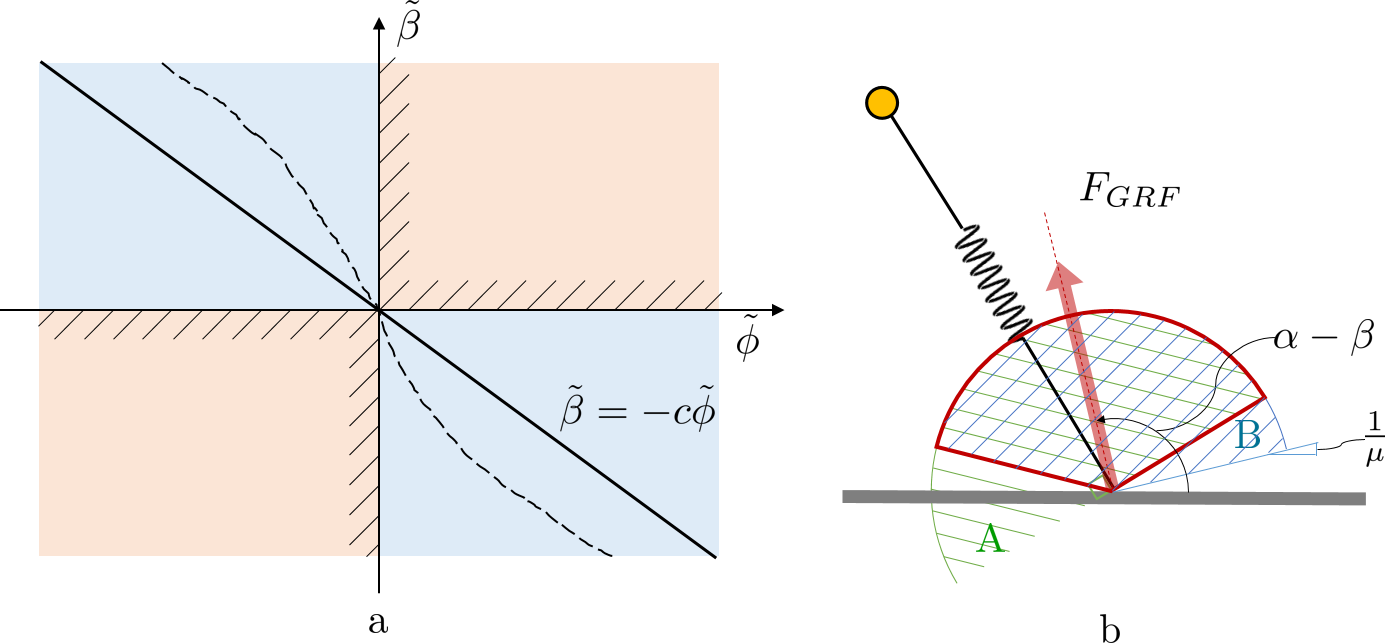}
        \caption{(a) Based on the analysis, we draw a basic rule for the direction of the GRF. $\tilde{\beta}$ can be any function of $\tilde{\phi}$ within 'Righting moment' region (dashed), but the simplest linear relationship (solid) is presented in this figure. (b) The feasible direction of the GRF is depicted (solid arc). Region A represents that the GRF redirection is accomplished using tangential force from the hip moment. Region B represents an arbitrary region that the user would define. One possible candidate is an estimated friction cone as depicted in the figure. }
    \label{fig3: possible controller}
%    }}
\end{figure}

Drawn upon the concepts, as shown in the above, a proper GRF redirecting controller should aim $\tilde{\phi}\cdot\tilde{\beta} < 0$, the region without shade in Fig. \ref{fig3: possible controller} (a). We would like to argue that if the generated GRF by a suitable redirecting controller satisfies this condition, any form of control would work. The simplest form we can propose is a redirecting controller having linear relationship as follows. 
\begin{align}
\label{eqn: new rule}
\tilde{\beta} = - c  \tilde{\phi}, \tilde{\phi} \in  [-\frac{\pi}{2}, \frac{\pi}{2}]
\end{align}
where $c$ is a positive gain which should be properly designed. 

It is worth noting that the speed and the direction of motion of the planar robot model should also be considered in determining the direction of the GRFs, we end up with the following control form for force direction control. 
\begin{equation}
\label{eqn: new rule vel}
\tilde{\beta} = - c  \tilde{\phi} - d\dot{\tilde{\phi}}, 
\end{equation}
where d is another positive gain, and it should be properly designed. Corresponding hip torques will be computed to redirect axial leg force to the desired direction with respect to CoM. The angle between the actual leg and the virtual line from foot to CoM can be computed from geometric information as well; denoting this angle as $\angle$, $\tan \beta = \tan (\tilde{\beta}+\angle)$ can be easily calculated.

\subsection{Feasible Direction of Foot-Ground Interaction Force}
The relation given in (\ref{eqn: new rule}) is expressed in terms of angles, therefore additional care is required for the controller. First of all, as the hip torque generates tangential force in $\tan \beta$, $\beta \in (-\frac{\pi}{2}, \frac{\pi}{2})$. if not, the direction of tangential force will be opposite the desired one. This is represented by the region A in Fig. \ref{fig3: possible controller} (b). Region A can be represented mathematically as follows
\begin{align}
A := \{\beta \in \mathbb{R} | -\frac{\pi}{2}<\beta < \frac{\pi}{2} \}
\end{align}

 Another thing we can consider is the range of the force in absolute reference. For example, we can consider friction cone. Although exact friction cone is hard to compute and thus not must thing, it improved the simulation result a lot. One can estimate the friction coefficient $\hat{\mu}$ and restraint controller generating force beyond friction cone. In this case, region B can be represented mathematically as follows. 
\begin{align}
B:=\{ \beta \in \mathbb{R}| \frac{\pi}{2}-\arctan\hat{\mu}<\alpha - \beta < \frac{\pi}{2}+\arctan\hat{\mu} \}
\end{align}
 
Even though the friction cone is neglected, at least we should consider that the GRF should be unilateral force,  \textit{i.e.}, the ground cannot pull the robot. Therefore, at least the region B should satisfy the follows.
\begin{align}
B:=\{ \beta \in \mathbb{R}| 0<\alpha - \beta < \pi \}
\end{align}

The intersection of the region A and B represents the feasible direction that the GRF can take, and that the controller would generate, $A \cap B $. 
\subsection{Foot Placement}
We adopt the velocity-based leg adjustment (VBLA) for swing leg control in order for regulation of linear momentum, which has been shown to be effective in enhancing stability and robustness of a point-mass walking mode \cite{sharbafi2016vbla}. The VBLA determines the desired swing leg direction $\textbf{O}$ as follows,
\begin{equation}
\textbf{O} = \mu \textbf{V} + (1-\mu)\textbf{G},
\end{equation}
where $\textbf{V} = \dfrac{v}{\sqrt[]{gL_0}}$ and $\textbf{G} = \dfrac{\textbf{g}}{g}$ are non-dimensional CoM velocity and gravitational acceleration. $\mu$ is the VBLA parameter to be determined properly. Swing leg touchdown angle $\alpha_0$ is obtained from the angle of the vector $\text{O}$. 
\subsection{Simulation Result of the Force Direction Controller}
\begin{figure}[ht!]
    \centering
    \includegraphics[width=1\textwidth,height=1.8\textheight,keepaspectratio]{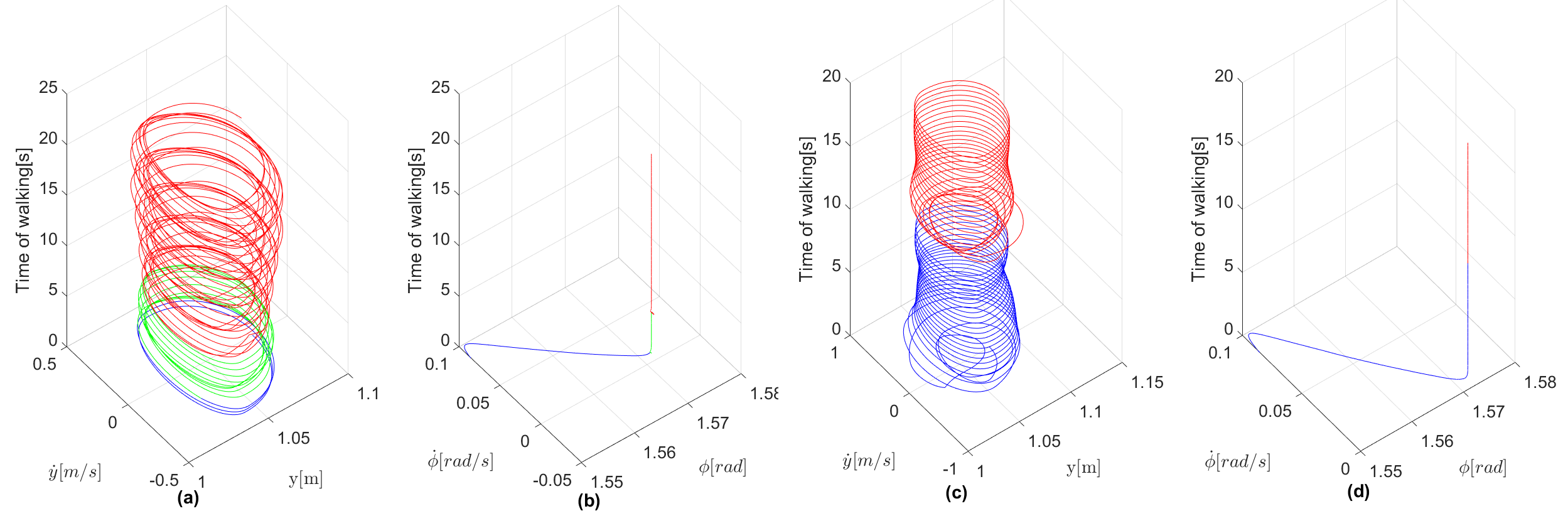}
    \caption{ Simulation result with a planar bipedal walking model with the proposed method on its sagittal plane. The response of the system with disturbance forces ((a) and (b)), and with rough terrain ((c) and (d), the maximum height of terrains \(h_{max} = 2cm\)). The phases plot (c) and (d) are changed color when the rough terrain is diminished at \(t = 10s\), while in (a), (b), the phases plot are changed color when external forces is applied at \(t=1[s]\) and $t=5[s]$.}
    \label{FDC}
\end{figure}
The dynamics of the bipedal walking model with trunk and compliant legs is simulated for 20 seconds, of which control parameters are listed in Table II. In Fig. \ref{FDC}, the phases plot of the center of mass vertical (y) motion and pitch ($\phi$) of the model are presented under disturbances forces ( (a) and (b)) and under terrain surface ((c) and (d)). In Fig. \ref{FDC} (a) and (b), external force disturbance of $F_{dist} = [30,100]^T[N]$ is applied to the right foot of the model during 0.3 seconds. The proposed method quickly stabilize the motion of pitch to its steady state motion, whereas the motion of CoM vertical oscillates in a small range. The method rejects this errors quite good in pitch motion. When the robot walks on the height map,  Fig. \ref{FDC} (c) and (d) shows the similar trend, The pitch motion is immediately stabilized, and at the same time, the translational motion is indirectly stabilized by the self-stabilizing property of a mass-spring walking system, as shown in the figure with solid red lines. When comparing with previous the combined control strategy, the force direction controller can give as much as performance without any precomputed trajectories and optimal control (DLQR). Therefore, we hope that method can open the door to build and control a real bipedal robotic platform based on the simple model and force direction controller. In the next chapter, we aim to apply the model to rigid-body articulated robotic model and develop a detailed control algorithm.
\section{Summary}
This chapter focused on control strategies of the BTSLIP model and provided two solutions for small terrain ground and disturbance forces in walking motion. 

Firstly, we proposed the combined control strategy by implementing the LQR control for adjusting the VPP location and feedback linearization control for legs stiffness. These results show that with the adaptability of the VPP position and the adjusting leg stiffness not only improves the stability but also for high robustness and fast disturbance rejection. 

Secondly, nature seems to take advantage of the attractive properties of mechanical templates and to facilitate stable gait without the need for precise trajectory planning. Therefore, without precomputed trajectories, We propose a force direction control method to regulate trunk motion while walking. We validate that three components of controls including spring leg, proposed force direction control, and proper foot placement realize robust walking of a reduced order model with respect to force disturbances. 

In the next chapter, we will create the controller for the articulated based on the BTSLIP model and second control method. 
\chapter{Control Strategy for the 5-link robot model based on Task-Space Control}
\chaptermark{Hierarchical Control Strategy for the robot model}
\label{chapter:chapter03}
\section{Introduction}
In Chapter 2, we validate that three components of controls including spring leg, proposed force direction control, and proper foot placement realize robust walking of a reduced order model on force disturbances. In this chapter, we will show how to use force direction controller with the simple model to ``guide'' the 5-link robot model by hierarchical control.  In general, hierarchical control strategy consists of two layers. In the first layer (higher level controller), a set of reference dynamic features (we used the leg forces profile) is generated from the simple template model based on certain control laws or optimization criteria. Then the second layer (lower level controller), these dynamic reference features are tracked by task space controller. On the other words, the 5-link robot model is controlled to have two virtual springy legs similar to the BTSLIP model. The desired GRF is computed to stabilize the trunk based on the force direction control. Besides, the proportional-derivative
The controller is applied to the virtual spring of swing leg. Then, we use the task space control to drive the 5-link robot to follow these sets of the target.

There are different methods to solve the task-space control (Operation Space Control - OSC). Khatib \cite{khatib1987unified} first proposed the operation space formulation to compute joint space actuator torques while considering task space dynamics. Later, Sentis and Khatib improved the OSC to deal with the underactuated and constrained dynamics of a humanoid robot \cite{sentis2005control}. Park and Khatib considered the control of contact forces of humanoid robots using OSC in \cite{park2006contact}. On the other hand, Aghili \cite{aghili2005unified} presented a method to simplify the computation of both direct and inverse dynamics for constrained systems using orthogonal projection, which is exploited by Mistry and Righetti \cite{mistry2012operational} who applied the method for the control of legged robots. Pratt et al. \cite{pratt2001virtual} proposed the intuitive control framework named as a Virtual Model control. In this structure, he used the simple Jacobian transpose to map the forces of virtual component (springs) into joint torques, this method neglect the Coriolis forces of the system dynamics. In this chapter, we present two approaches to task space control. In the first proposed task space control, we use the OSC formulation from \cite{mistry2012operational} for constrained and underactuated system to compute the joint torques of the robot model. The second method of task space control is from the simple Jacobian Transpose \cite{pratt2001virtual}. In this method, we do not intend to force the dynamics of the five-link robot model to exactly follow those of the reduced order model. The goal is rather to exploit the fundamental properties verified with a reduced order model to control a system with substantially more complex dynamics. In this perspective, we show that a simple form of the controller can still show adequate performance for walking, without using model-based controls or precomputed trajectory. 

\begin{figure}[ht]
    \centering
    \includegraphics[width=0.8\textwidth,height=0.9\textheight,keepaspectratio]{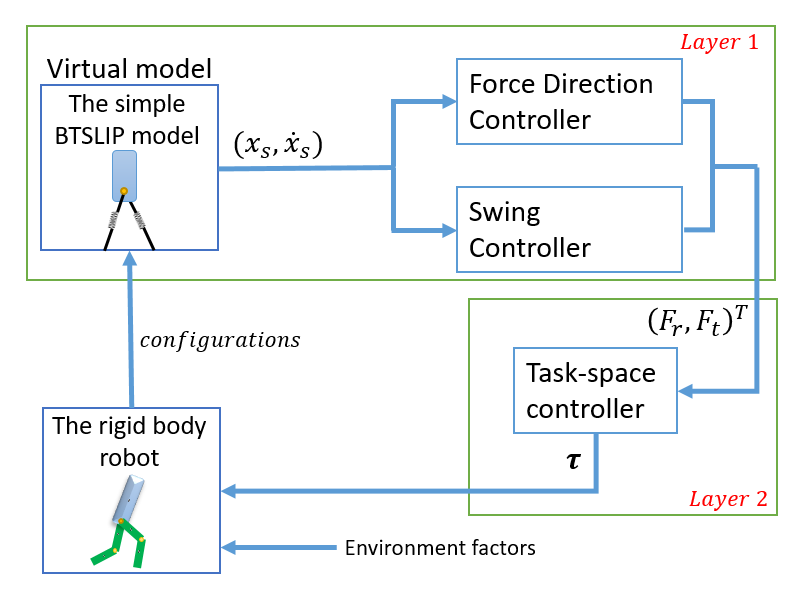}
    \caption{Block diagram of the simple-model-guided task space control system. We create virtual springs between the hip joint and foot-end of two legs, as utilized in the simple models. As in the simple virtual model, each leg controller has swing and stance mode which correspondences to the force direction controller and swing controller in layer 1. The controller in Layer 1 proposed forces at foot-end, which is converted into joint torques via Jacobian Transpose or Operation Space Control. In the simulation, we also test the walking motion under disturbance forces and rough terrain.}
    \label{fig:blockdiagram}
\end{figure}

Fig. \ref{fig:blockdiagram} provides an overview of the ``simple-model-guided task space control framework". The details in this Figure will be discussed in more detail in the following sections.

The remainder of the chapter is organized as follows. In Section 3.2, we provide brief descriptions of the robot model. Section 3.3 presents the control strategy for the robot model. Section 4 follows with simulation results which demonstrate the effectiveness of the control strategy. Finally, we conclude the paper with final remarks section 5.
\begin{table}[h]
\normalsize
\caption{Model parameters}
\label{table_example}
\begin{center}
\begin{tabular}{| m{3cm} | m{1.1cm}| m{1.1cm} | m{1.1cm} |}
\hline    
Model parameters & Trunk (t) & Femurs (f) & Shins (s) \\
\hline
\(m_{*}\) (\(kg\)) & 12.5 & 0.7 & 0.7  \\
\(L_{*}\)(\(m\)) & 0.42 & 0.19 & 0.19\\
\(J_*\) (\(kg m^2\)) & 0.23 & 0.0045 & 0.0045 \\
\(c_*\)(\(m\)) & 0.21 & 0.13 & 0.13 \\
\hline
\end{tabular}
\end{center}
\end{table}
\section{Robot model}
The robot model in this study  consists of five links as depicted in Fig. \ref{fig:first}: a trunk and two identical lower limbs with each limb having a femur and a shin; moreover, all links have mass and are connected by revolute joints. All of the joints are considered only rotating in the sagittal plane. The walking cycle consists of two phases:  the single support (SS, the phase in which only the stance leg is touching the ground) and the double support (DS,  both legs are in contact with the ground). The multi-rigid-body contact model of \cite{hurmuzlu1994rigid} is assumed as the underactuated revolute joint with constraints of uni-laterality (no lift-off) and no slipping between leg and walking surface in SS. That collapses the double support phase to an instant time (impacts) and allows a discontinuity in the velocity of states. 

To describe the robot system,  we use the Lagrangian formulation and the assumption of rigid bodies. The model parameters are given in Table I. 
  
\begin{figure}[ht]
    \centering
    \includegraphics[width=0.8\textwidth,height=0.9\textheight,keepaspectratio]{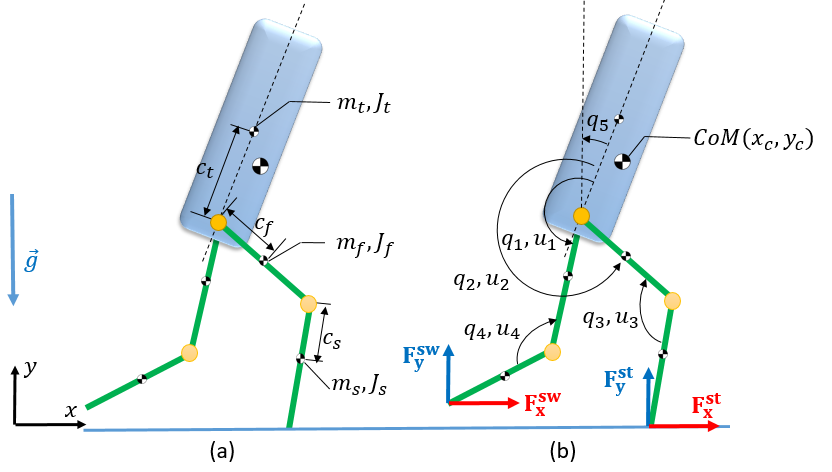}
    \caption{The robot model parameters (a). The robot model coordinates follow the convention such that angles and torques are defined positive in counter-clockwise direction (b). \(q_5\) is the absolute trunk angle. The position of the floating trunk are denoted as \([x_c,y_c]\)). \(\) in Cartesian coordinates. }
    \label{fig:first}
\end{figure}
\subsection{Single Support}
In the single support, legs of the robot are marked as stance and swing. With 5-link, the dynamic model consists of seven DOF, where four actuated DOF associated with the joint coordinates, two underactuated DOF associated with the horizontal and the vertical displacements of the Center of Mass (CoM), and last one underactuated DOF associated with the orientation of the robot in a sagittal plane. Thus, the generalized coordinates of the system (\(q_e\)) can be combined by two subsets \(q\) and \(r\).
\begin{equation*}
\label{eqn: coordinates}
q_e := [q,r]^T 
\end{equation*}
where \(q:= [q_1,q_2,q_3,q_4,q_5]\) encapsulates the joint coordinates, \(q_5\) is the underactuated DOF which is associated with orientation of the robot. $q_1$ and $q_3$ describe the configuration of the swing leg, while $q_2$ and $q_4$ are for the stance leg. Besides, \(r:=[x_c,y_c] \in \mathbb{R}^2\) is the Cartesian coordinates of the CoM of robot. The second order dynamical model follows from Lagrange's equation \cite{goldstein1965classical}, \cite{spong2006robot}.
\begin{equation}
\label{eqn:EOM}
\begin{aligned}
& D_e(q_e)\ddot{q}_e + C(q_e,\dot{q_e}) + G_e(q_e) = B_e \tau + J_e^{st}(q_e)^TF^{st} \\
\end{aligned}
\end{equation}
where \(D_e(q_e) \in \mathbb{R}^{7 \times 7}\) is the symmetric and positive definite inertia matrix, \(C(q_e,\dot{q}_e) \in \mathbb{R}^7\) summarizes the centrifugal and Coriolis terms, \(G(q) \in \mathbb{R}^7\) is the vector containing gravity terms. \(B_e \in \mathbb{R}^{7 \times 7}\) is the input matrix (i.e. \(B_e = \begin{bmatrix}
    I_{4\times 4}   & 0   \\
   0 & 0      
\end{bmatrix}
\) with \(I_{4\times 4}\) is the identity matrix) and \(\tau:= [u_1,...,u_4,0,0,0]^T \in \mathbb{R}^7\) includes the joint torques applied at the joints of the robot. \(J_e^{st} := \dfrac{\partial p^{st}}{\partial q_e} \in \mathbb{R}^{2 \times 7}\), \(p^{st}\) is the Cartesian coordinates of stance foot. Also,  \(F^{st} := [F_x^{st},F_y^{st}] \in \mathbb{R}^2\) is constraint force which corresponds to  the ground reaction force at the stance leg end.

We assume that a stance foot on the ground is only allowed to perform a pure rolling motion with neither slipping nor sliding. This means we have the constraint below:
\begin{equation}
\label{eqn:constraint}
\begin{aligned}
& \dot{p}^{st} = J_e^{st}\dot{q}_e = 0 \\
& \ddot{p}^{st} = J_e^{st}\ddot{q}_e + \dot{J}_e^{st}\dot{q}_e = 0
\end{aligned}
\end{equation}
\subsection{Impact}
An impact takes place when the swing leg touches the ground as a collision between two rigid bodies \cite{hurmuzlu1994rigid}, \cite{hyun2014high}. The impact is assumed to be inelastic and instantaneous. When the impact occurs, the leg previously pinned on the ground (i.e., the stance leg) loses contact, and the roles of legs are switched. Then, the impact model which meets the standard hypotheses in \cite{grizzle2001asymptotically}, results in the impact equation which can be expressed by the following equation: 
\begin{equation}
\label{eqn:impact}
\begin{aligned}
& D_e(q_e)\dot{q}_e(t^{+}) - D_e(q_e)\dot{q_e}(t^{-}) = J_e^{sw}(q_e)^T \delta F^{sw} \\
& J_e^{sw}(q_e)\dot{q}_e(t^+) = 0
\end{aligned}
\end{equation}
where \(J_e^{sw} := \dfrac{\partial p^{sw}}{\partial q_e} \in \mathbb{R}^{2 \times 7}\), \(p^{sw}\) is the Cartesian coordinates of swing foot and  \(\delta F^{sw}:= \int_{t^-}^{t^+} F^{sw}(\tau)d\tau \) is the impulsive force at impact point. The superscript \(+\) and \(-\) denotes the the post-impact and pre-impact of system, respectively. The solution of (\ref{eqn:impact}) can be more clearly given by the following equation: 
\begin{equation}
\label{eqn:sort}
\begin{bmatrix}
    \dot{q}_e(t^+)      \\
    \delta F^{sw}      
\end{bmatrix}
= \Delta(q_e)
\begin{bmatrix}
    D_e(q_e)\dot{q}_e(t^-)      \\
    0      
\end{bmatrix} 
\end{equation}
where 
\(
\Delta(q_e)
= 
\begin{bmatrix}
    D_e(q_e) & -J_e^{sw}(q_e)^T      \\
    J_e^{sw}(q_e) & 0      
\end{bmatrix} ^{-1}
\)
. Instead of introducing additional equations of motion for the next single support with new stance leg, the transform of the coordinates of the robot is done by relabeling matrix \(R \in \mathbb{R}^{7 \times 7}\). Formally, the combination with  (\ref{eqn:sort}),  the state of the system after impact can be expressed as:
\begin{equation}
\label{eqn:finalrigid}
\begin{aligned}
& q_e(t^+) = Rq_e(t^-) \\
& \dot{q}_e(t^+) = R\Delta_{11}(q_e(t^-))\dot{q}_e(t^-),
\end{aligned}
\end{equation}
where $\Delta_{11}$ is the first \(7 \times 7\) matrix of \(\Delta(q_e)\)
\subsection{Controller Modification for the Planar Five Link Model} 
In this section, the control strategy for robot model is described. We first derive desired forces to be exerted at foot of the model for both stance and swing legs. In particular, the desired force of the stance leg is determined based on simple models; a virtual spring is created to connect hip and foot of the robot model, as in the Bipedal Trunk Spring Loaded Inverted Pendulum (BTSLIP) model. A proper force direction control rule is adopted from \cite{lee2017biped} in order for bipedal walking with upright trunk. Besides, simple PD control is implemented for swing leg. Finally, the operational space control (OSC) is used to map the desired force to joint actuators. The control parameters are presented in Table \ref{table:control}.
\subsubsection{The stance leg control} 
\begin{table}[h!]
\caption{Control parameters for the the 5-link robot model}
\label{table:control}
\begin{center}
\begin{tabular}{|c|c|c|c|c|}
\hline
Parameter & Meaning  & Value [unit]\\
\hline
$k$ & spring stiffness for virtual leg & 7500 [$N/m$]\\
$c$ & position-proportional gain & 10 [$\cdot$] \\
$c_{sw}$ & position-proportional gain & 10 [$\cdot$] \\
$d$ & velocity-proportional gain  & 1 [$s$]\\
$\mu$ & coefficient value for VBLA & 0.5 [.] \\
$L_0$ & the rest length of virtual leg & 0.37 [$m$]\\ 
$k_d$ & virtual spring damping &100 [$N.s/m$]\\
\hline
\end{tabular}
\end{center}
\end{table}
\begin{figure}[ht]
    \centering
    \includegraphics[width=0.9\textwidth,height=0.85\textheight,keepaspectratio]{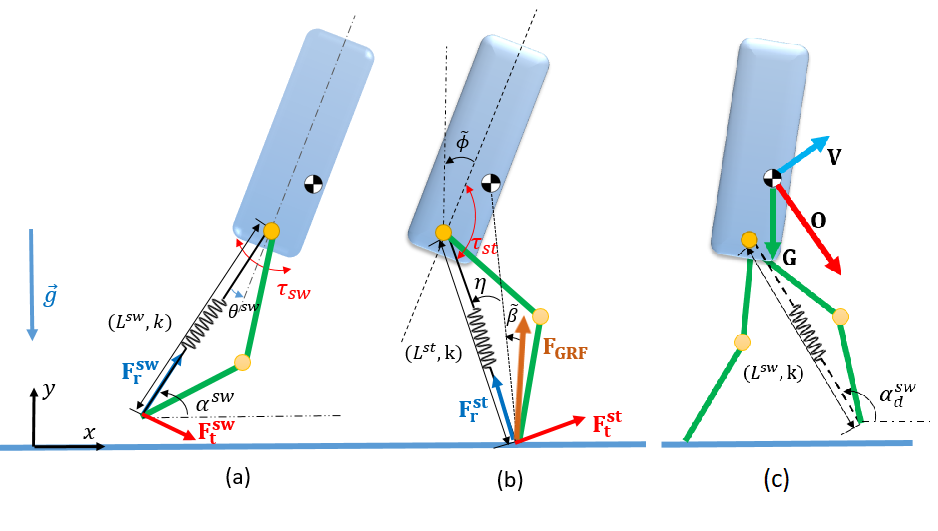}
    \caption{The robot model is considered to have virtual compliant legs. Two virtual spring is connected from hip to each foot with the same spring stiffness, \(k\). The control variables for swing leg (a). And the control variables for stance leg (b). Velocity-based leg adjustment \cite{sharbafi2016vbla} (c).}    
\label{fig:control}
\end{figure}
Fig. \ref{fig:first} shows that the 5-link model will have virtual springs from hip to each foot. We embedded the controller of the BTSLIP model \cite{maus2010upright}, \cite{lee2017biped} in order to calculate the virtual axial force and tangential force that is necessary to maintain the upright trunk in walking for the stance leg:
\begin{equation}
\begin{aligned}
&F_r^{st} = k (L_0 - L^{st}) + k_d\dot{L}^{st}\\
&F_t^{st} = F_r \tan \beta
\end{aligned}
\end{equation}
where \(k\) and \(L_0\) are the spring stiffness and the rest length of virtual leg, respectively.  \(L^{st} = | p^{st}_f - p_h|\) is current leg length, where \(p^{st}\) and \(p_h\) are the position of stance feet and hip in global frame, respectively. $\beta = \eta + \tilde{\beta}$ is the angle between \(\textbf{F}_{GRF}\) and the virtual leg, and $\eta$ is the angle between the vector from stance foot to CoM and virtual leg. We added the damping component in computing of the axial force. Note that we set up the lower limit of \(F_r\) to zero. 

The ground reaction force vector \(\textbf{F}_{GRF}\) is the combination of the virtual spring force and reaction force : $\textbf{F}_{GRF} = \textbf{F}_r^{st} + \textbf{F}_t^{st}$. The direction of $\textbf{F}_{GRF}$ is controlled by $\tilde{\beta}$. We take into account the speed and direction of the trunk in determining the direction of \(\textbf{F}_{GRF}\) as follows: 
\begin{equation}
\begin{aligned}
&\tilde{\beta} = -c\tilde{\phi} - d\dot{\tilde{\phi}},
\end{aligned}
\end{equation}
where $c$ and $d$ are the control parameters, as shown in Table \ref{table:control}. $\tilde{\phi}$ is the pitch of the trunk.
Besides, the desired GRF vector is decomposed into horizontal \(F_x^{st}\) and vertical component \(F_y^{st}\) in the global frame using an angle \(\theta_{p}^{st}\)(\(= \alpha^{st} - \beta\)): 
\begin{equation}
\begin{bmatrix}
     F_x^{st}      \\
     F_y^{st}      
\end{bmatrix}
= \begin{bmatrix}
    F_{GRF}\sin(\theta_{p}^{st})     \\
    F_{GRF}\cos(\theta_{p}^{st})   
\end{bmatrix} ,
\end{equation} 
where $F_{GRF} = ||\textbf{F}_{GRF}||$ is the magnitude of the GRF vector.
\subsubsection{The swing leg control} 
We use the method velocity-based leg adjustment (VBLA \cite{sharbafi2016vbla}) for swing leg control, to define the touchdown angle for swing leg, as shown in Fig. \ref{fig:control}(c). The VBLA determines desired swing leg orientation as follows, 
\begin{equation}
\textbf{O} = \mu \textbf{V} + (1-\mu)\textbf{G},
\end{equation}
where \(\textbf{V} = \dfrac{\textbf{v}}{\sqrt{gL_{sw}}}\) and \(\textbf{G} = \dfrac{\textbf{g}}{|g|}\) are non-dimensionalized CoM velocity and gravitational acceleration, respectively. $\textbf{v}$ and $\textbf{g}$ are the vector of CoM velocity and gravity. The angle between the desired vector \(\textbf{O}\) and the ground is determined as the desired touch down angle \(\alpha_d^{sw}\). By defining the virtual leg length trajectory for swing leg \(L_d\), and combining with the desired touch down angle from VBLA, the simple form of proportional - derivative control can be applied for swing leg control: 
\begin{equation}
\begin{aligned}
& F_r^{sw} = k(L_d - L^{sw}) + k_d \dot{L}^{sw} \\
& \tau_{sw} = c_{sw}(\alpha_d^{sw} - \alpha^{sw}) + \dot{\theta}^{sw},
\end{aligned}
\end{equation}
where \(L_{sw} = |p^{sw}_f - p_h|\) is the length of the virtual swing leg, \(p^{sw}_f\) is the position of the swing foot in global frame, and  \(\alpha^{sw}\) is the angle created by the virtual swing leg and the horizontal axis. \(c, k_d\) and \(c_{sw}\) are the control parameters defined in Table \ref{table:control}. The virtual leg length can be retracted to prevent the scuffing of leg with the walking surface by controlling the desired swing leg length \(L_d\): 
\begin{equation}
\begin{aligned}
\left \{
\begin{array}{ll}
 L_d = \dfrac{4}{5} L_0 \: \text{if} \: -\dfrac{\pi}{9}\leq (\alpha^{sw} - \dfrac{\pi}{2}) \leq \dfrac{\pi}{18} \\
 L_d = L_0 \: \: \: \: \text{if} \:\: (\alpha^{sw} - \dfrac{\pi}{2}) > \dfrac{\pi}{18} 
 \end{array}
\right.
\end{aligned}
\end{equation}
 By applying the torque \(\tau_{sw}\), it is equivalent to create the tangential force along with the virtual swing leg \(F_t^{sw} = \dfrac{\tau_{sw}}{L_{sw}}\), as shown in  Fig. \ref{fig:control}(b). The combination of tangential force and normal force creates the total desired force at end-effector of swing leg: \(\textbf{F}_{sw} = \textbf{F}_t^{sw} + \textbf{F}_r^{sw}\). Also, the end-effector force vector of swing leg is projected in the global frame with angle \(\theta^{sw}_{p}\):
\begin{equation}
\begin{aligned}
\begin{bmatrix}
     F_x^{sw}      \\
     F_y^{sw}      
\end{bmatrix}
= \begin{bmatrix}
    F^{sw}\sin(\theta_{p}^{sw})     \\
    F^{sw}\cos(\theta_{p}^{sw})   
\end{bmatrix} 
\end{aligned},
\end{equation}
where $F^{sw} = ||\textbf{F}^{sw}||$ and \(\theta^{sw}\) is the angle between \(\textbf{F}_{sw}\) and horizontal line. 
\subsection{Mapping the desired end-effector force with the actual joint torques of robot}
\subsubsection{Operational Space Control for Underactuated and Constrained system}
We consider that \(J^{st}\) and \(J^{sw}\) are the Jacobian of the vector along with the virtual swing leg and stance leg in the global frame, respectively:
\begin{equation}
\begin{aligned}
& J^{st} = \dfrac{\partial(p_h - p^{st}_f) }{\partial q_e}; \:
& J^{sw} = \dfrac{\partial(p_h - p^{sw}_f) }{\partial q_e},
\end{aligned}
\end{equation}
where \(J^{st} \in \mathbb{R}^{2 \times 7}\) and \(J^{sw} \in \mathbb{R}^{2 \times 7}\). In more compact form, we assign \(J = [J^{st};J^{sw}]\) and \(F = [F_x^{st},F_y^{st},F_x^{sw},F_y^{sw}]^T\). 

In order to compute joint torques, we adopt the OSC introduced in \cite{mistry2012operational} which is formulated for constrained and underactuated system. This method exploits the null space motion and constraint forces of the system to accomplish the tasks in operational space while minimizing the magnitude of null-space torques. The control inputs can be computed as
\begin{equation}
\label{constrain} 
\tau = J^T F + N \tau_0,
\end{equation}
where \(N = I - J^T J^{T^\#}\), \(J^{T^\#} = (JM_cPJ^T)^{-1}M_c^{-1}P\),  and \(\tau_0\) is an arbitrary null space torque. P is an orthogonal projection operator which is readily computable from the constraint Jacobian \(P = I - (J_e^{st})^+ J_e^{st}\) \cite{aghili2005unified} (+ indicates the Moore-Penrose pseudoinverse) and \(M_c = P D_e + I -P\) is invertible.

In order for minimizing the value of \(||\tau_0||\) (Euclidean norm of vector \(\tau_0\)), the null-space torque $\tau_0$ is formulated as 
\begin{equation}
\label{tau0}
\tau_0 = -[(I - B_e)N]^+ (I-B_e)J^T F,
\end{equation}
where \(I\) is the identity matrix. By substituting (\ref{tau0}) into (\ref{constrain}), we can  write the control equation as in \cite{mistry2012operational} 
\begin{equation}
\label{20}
\tau = (I-N[(I-B_e)N]^+)J^T F.
\end{equation}
\subsubsection{Polar Jacobian Transpose method}
\begin{figure}[ht]
    \centering
    \includegraphics[width=0.6\textwidth,height=0.7\textheight,keepaspectratio]{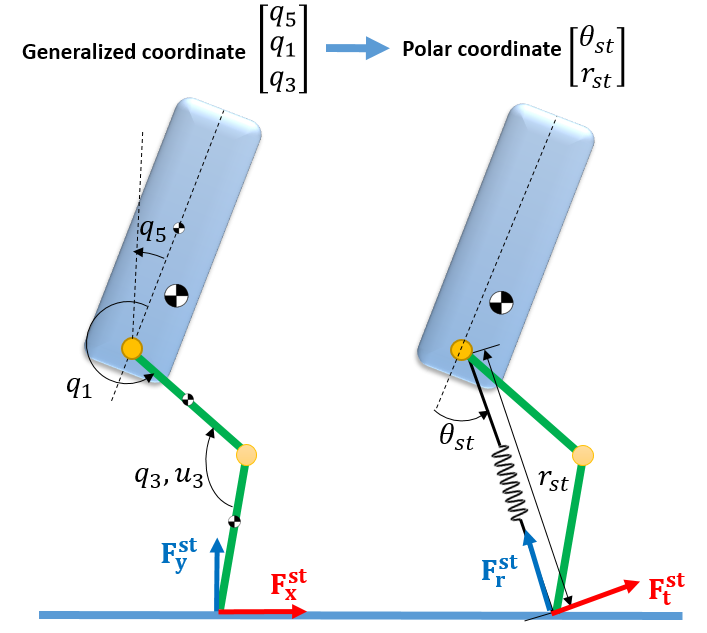}
    \caption{Polar coordinate transformation for stance leg. The similarity will be applied for the estimation in swing leg.}    
\label{fig:control}
\end{figure}
Fig. shows the transformation between generalized coordinate of stance leg to polar coordinate, the similar calculation is applied for the swing leg. The stance leg control and swing leg control measure the forces at foot end with respect to a hip as $F_{polar} = [F_r^{st},F_t^{st},F_r^{sw},F_t^{sw}]^T$. The Jacobian from hip to foot end of each leg is transformed to polar coordinate and multiplied with the set of forces at foot end to calculate torque commands for each joint as: 
\begin{equation}
\tau=J^T_{polar}F_{polar}
\end{equation}
\subsection{Simulation results}
\begin{figure*}[h]
    \centering
    \includegraphics[width=\textwidth,height=6.5cm]{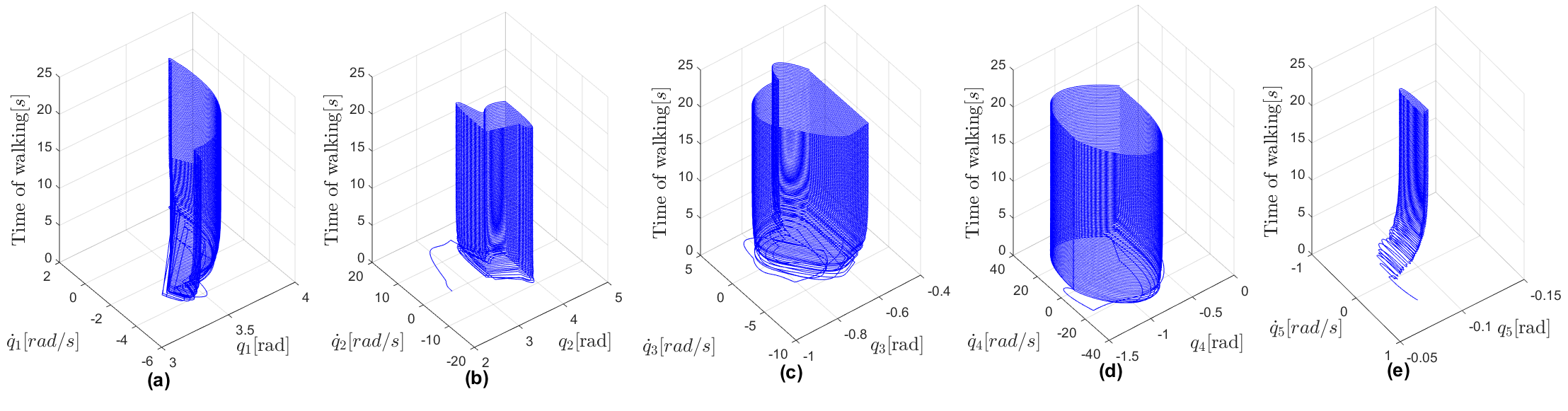}
    \caption{Simulation result of the planar 5-link robot model walking 100 steps with proposed control method. The model begins with a random initial condition and converges to its periodic trajectory within ten steps. The phase portrait of an angle from trunk to stance leg (a) and to swing leg (b). The phase plot of stance knee angle (c) and swing knee angle (d). The phase plot of absolute trunk angle (e).} 
\label{fig:result1}
\end{figure*}
To test the validity of the proposed control method with the 5-link rigid model, we use the dynamic simulation. The model parameters and control parameters are listed in Table 3.1 and Table 3.2, respectively. The 5-link robot dynamics and the proposed controller are illustrated in Section 3.2. To be a more realistic simulation, we include a small amount of viscous friction at the joints. 
\subsubsection{The simulation results of Operational Space Control}
We use ode45 integrator in Matlab to create the dynamics of 5-links biped robot with impact assumptions in Section 3.2.2. 
To test the validity of the proposed control method with the 5-link rigid model, we use the dynamic simulation. The model parameters and control parameters are listed in Table I and Table II, respectively. The 5-link robot dynamics are illustrated in Section II, and the proposed control is presented in Section III. Besides, we include a small amount of viscous friction at the joints. 
\begin{figure}[ht!]
    \centering
    \includegraphics[width=0.8\textwidth,height=0.7\textheight,keepaspectratio]{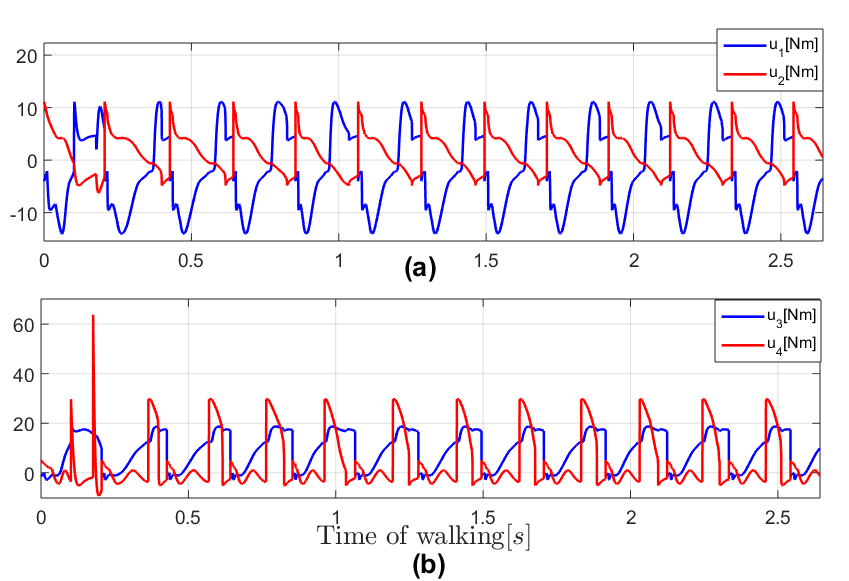}
    \caption{ The control inputs at hip of stance ($u_1$) and swing leg ($u_3$), respectively in (a) and the control inputs at knee of stance ($u_2$) and swing ($u_4$) in (b) over initial few steps (2.5 seconds).}  
\label{res:control}
\end{figure}

Fig. \ref{fig:result1} presents all the phase plots of the joint coordinates \(q\) with the initial condition 
\begin{equation*}
q_e=[3.7,2.6,-0.5,-0.4,-0.08,-2.4,-0.1,-1.3,-0.9,0.5]^T, 
\end{equation*}

(units are omitted). Note that the walking motion starts with the random initial condition, thus at the beginning of the simulation when time $t<2$, it has a lot of jerky motion, but after some steps, the proposed method quickly stabilizes the model to the steady-state motion. The absolute trunk angle is maintained the upright posture with small oscillation \([-0.095:-0.085]\)[rad]. Furthermore, after approximately 5 seconds, the model converges to its periodic trajectory. 

\begin{figure}[ht]
    \centering
    \includegraphics[width=\textwidth,height=7.5cm,keepaspectratio]{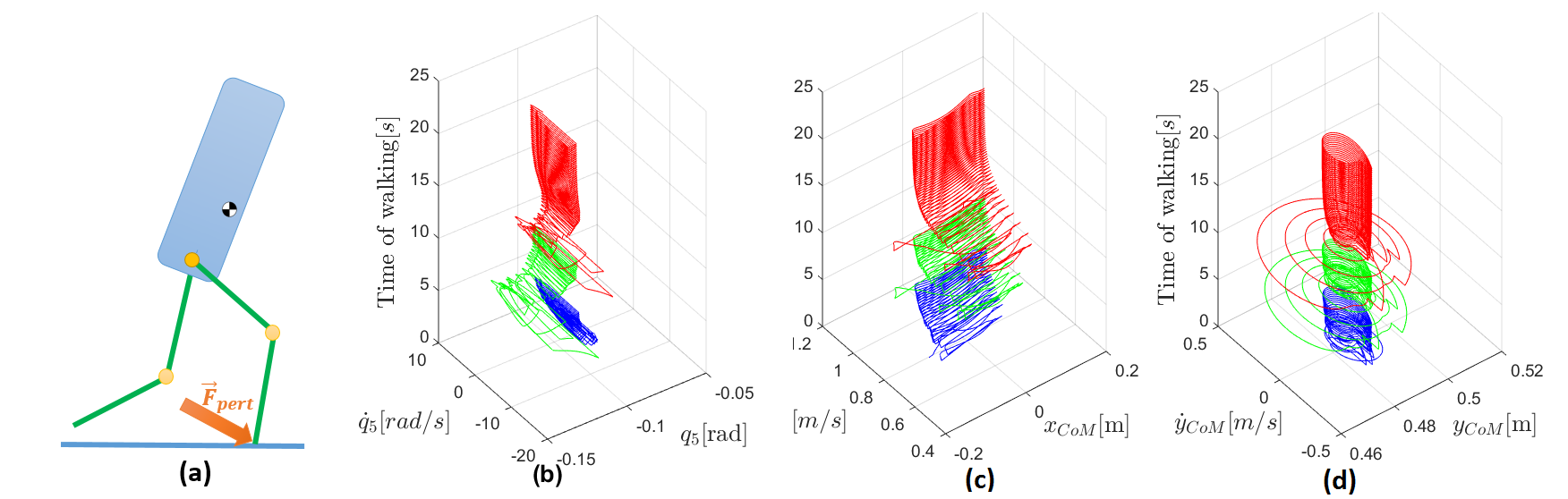}
    \caption{ The external  force disturbance is applied at the stance foot (a). The phases portrait of trunk angle (b), the horizontal position (c), and the vertical position (d) with the same initial condition with Fig. \ref{fig:result1}, but under the disturbances. The phases plot are changed color to green and red after the disturbances is applied at \(t = 5s\) and \(t = 10s\).}
\label{fig:pert}
\end{figure}
It can be seen in Fig. \ref{res:control}  that the control signals oscillate with large range at the very first step. After that, the steady state behavior of control inputs is attained with the range of hip torques control is about \([-15,11]^T\)[Nm] and \([-5,30]^T\)[Nm] for the knee torques. 

To further investigate the performance of the proposed control, the external force disturbance (\(F_{pert} = [5 \: 10]^T\)) is applied to the stance foot, as seen in Fig. \ref{fig:pert} (a). Fig. \ref{fig:pert} (b), (c), and (d) show the phase plots of absolute trunk angle ($q_5$), CoM horizontal($x_{CoM}$) and vertical position($y_{CoM}$). The force disturbances can affect the horizontal, vertical motion, and the rotation of the trunk. It is obvious that the proposed control is effective and induces gait stability in the 5-link robot model. The trunk is quickly stabilized after applying perturbation; the effects of disturbance are also rejected in horizontal and vertical motion. 
\begin{figure}[h!]
    \centering
    \includegraphics[width=0.65\textwidth,height=0.5\textheight,keepaspectratio]{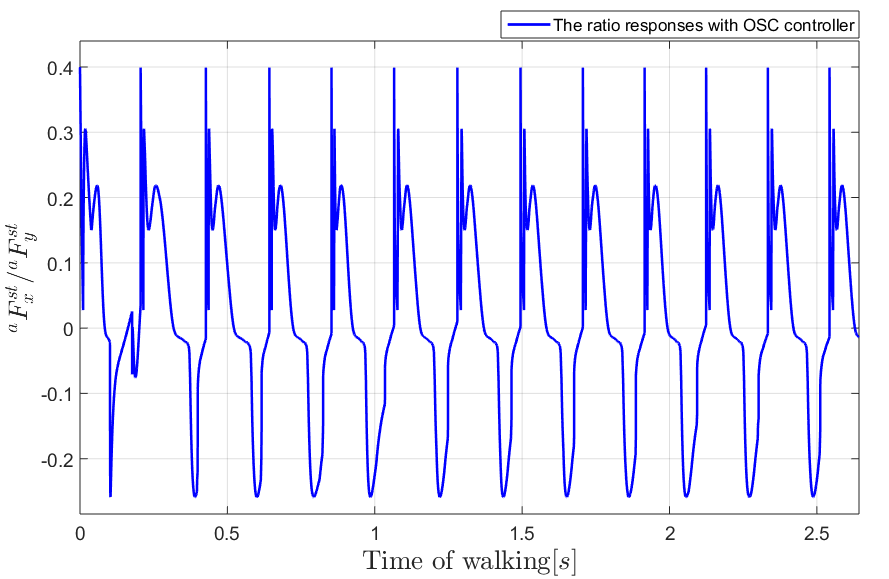}
    \caption{The ratio of two components of actual GRF (tangential force and normal force)}  
\label{fig:compare}
\end{figure}
Fig. \ref{fig:compare} shows that the assumptions we made for the simulation is not violated. In this figure, the ratio of the horizontal and vertical components of the actual GRF (\(^aF^{st} = [^aF_x^{st},^aF_y^{st}]^T\) - the superscript `a' means actual), which is a paramount notion of slippage in walking, are plotted for a few initial steps. By assuming the Column friction model, \(|^aF^{st}_{x}/^aF^{st}_y| \leq \mu_{fric}\) will guarantee the non-slippery condition of stance leg. It is notable that the model would not slip while walking in the environment with \(\mu_{fric} \geq 0.4 \), as presented in Fig. \ref{fig:compare}. 
\subsubsection{The simulation results with Polar Jacobian Transpose}
We use the Open Dynamic Engine platform which a physics simulation engine to test the performance of proposed control. To avoid the burden of computation cost and simplify the control strategy, we use the classical method Jacobian Transpose to compute the joint torques. 

\begin{figure*}[h!]
    \centering
    \includegraphics[width=11cm,height=7.5cm]{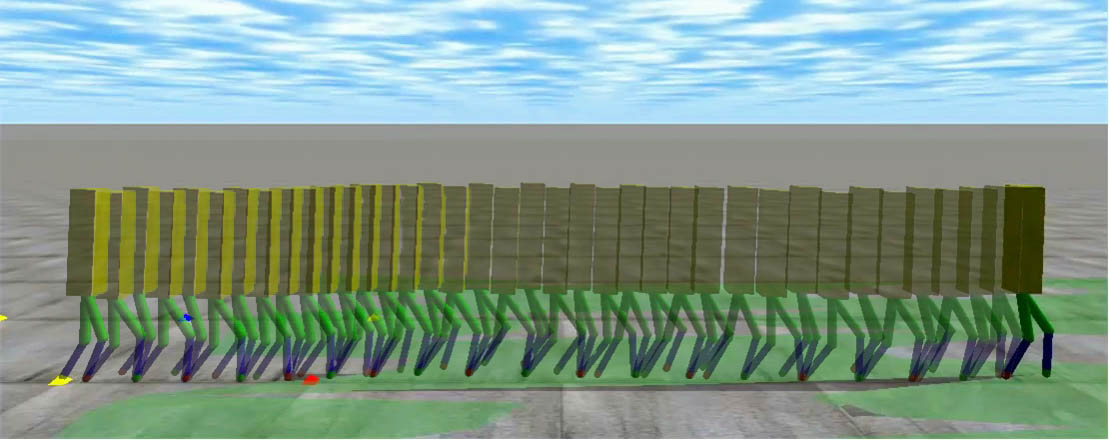}
    \caption{Snapshots of several steps in approximately 10 seconds on terrain ground.} 
\label{fig:snapshots}
\end{figure*}

To create the rough terrain in ODE, we use height map terrains function in ODE. The height level of each point in ground terrain map will be defined by formulation: $z(cm) = 1 + 5\sin(20y+20x)$, where $x$, $y$ and $z$ are the location of each point on the terrain map. More clearly, the height of each point in terrain map will be in the range $[0:6]cm$ which correspondences to $20\%$ of model leg length. The snapshots of the robot model on rough terrain are shown in Fig. \ref{fig:snapshots} in $10[s]$ with interval $\delta t = 0.001$. 

\begin{figure*}[h!]
    \centering
    \includegraphics[width=\textwidth,height=5cm]{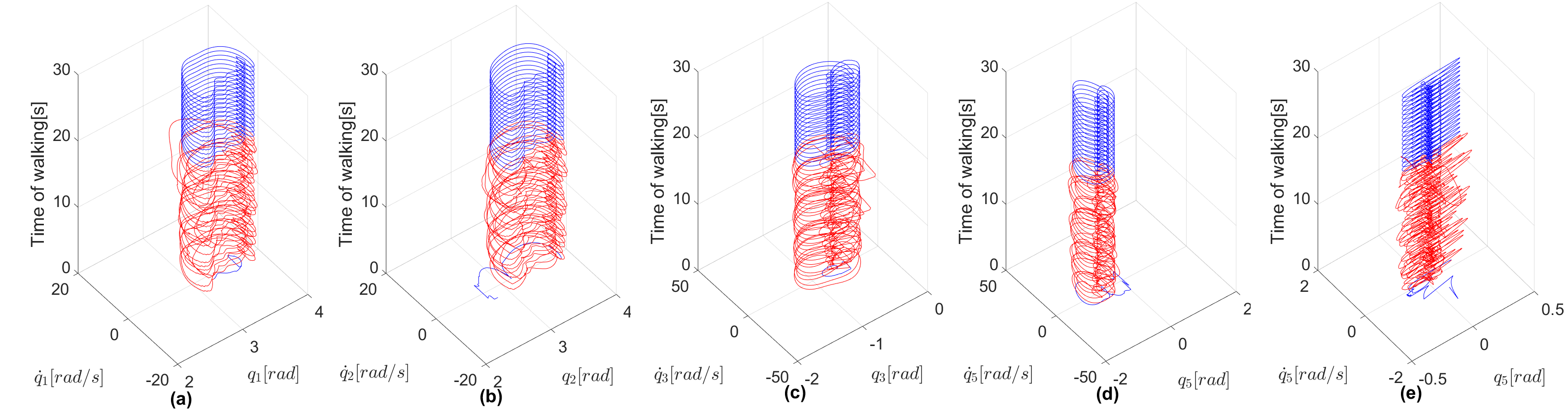}
    \caption{The model begins walking on the flat ground and quickly moves to the rough terrain. The phase portrait of an angle from trunk to the thigh of the leg 1 (a) and leg 2 (b). The phase plot of stance knee angle of the leg 1 (c) and the knee angle of the leg 2 (d). The
phase plot of absolute trunk angle (e). The phases plot are changed to red color when robot is in the rough terrain surface.} 
\label{fig:roughterrain_phaseplot}
\end{figure*}
\begin{figure}[h!]
    \centering
    \includegraphics[width=\textwidth,height=0.4\textheight,keepaspectratio]{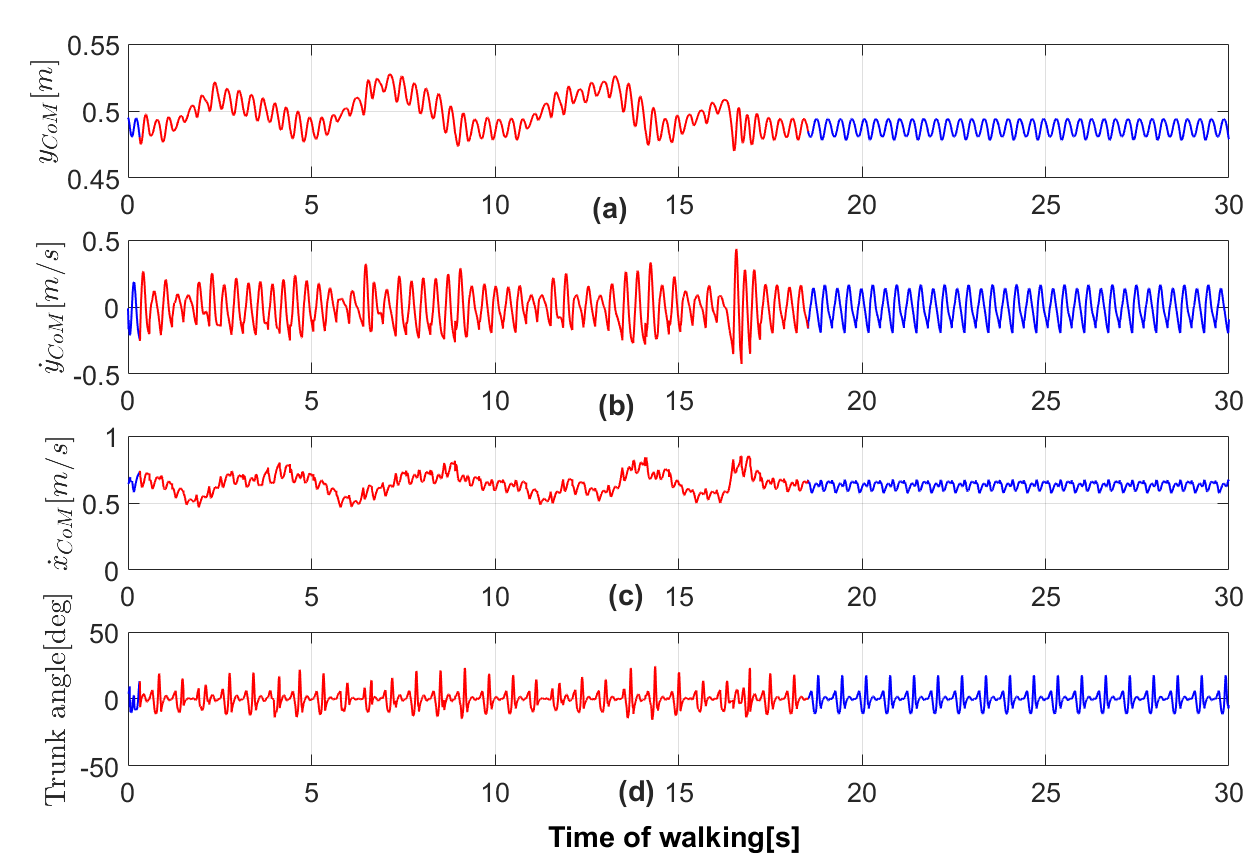}
    \caption{ Trajectories in 30 seconds of the 5-link robot model. The vertical position of the CoM (a). The velocity of vertical position of CoM (b). The forward velocity of the robot model (c). The pitch of trunk is plotted in (d)}
\label{fig:trajectory_terrin}
\end{figure}
\begin{figure}[h!]
    \centering
    \includegraphics[width=\textwidth,height=0.35\textheight,keepaspectratio]{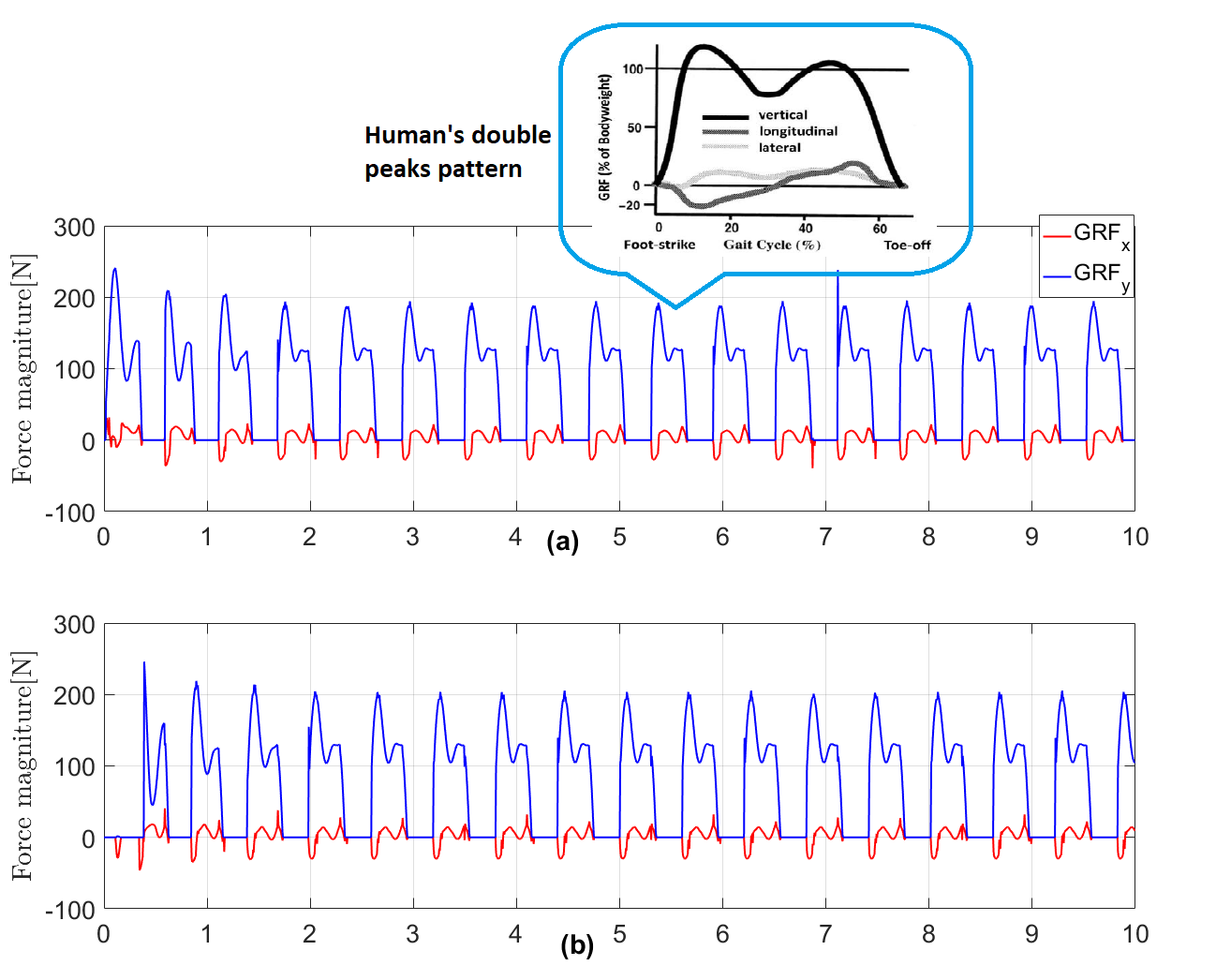}
    \caption{The measured ground reaction force profiles of the robot during walking. The forces can be very similar to the double peaks pattern of human ground reaction force which suggests that humans control their legs to act as linear springs during steady state walking and running.}
\label{fig:GRF}
\end{figure}
The graph of Fig. \ref{fig:roughterrain_phaseplot} shows the phase plots of joints angle. We have found that the controller constructs a stable limit cycle when walking on flat ground and although it loses periodicity, the robot model can maintain walking on rough ground as well. More clearly, in the graph, the jerk appears when robot walks on the rough terrain and quickly diminishes in flat ground. Fig. \ref{fig:trajectory_terrin} shows the trajectories of some states in 30 seconds. It is obvious that when the robot is on the rough surface, the springy leg tried to change the height of CoM to adapt. In the graph, the trajectories of vertical position and forward velocity oscillates while the trunk orientation is stable with a small range of oscillation. It is interesting that the algorithm does not contain any explicit speed control mechanism, yet the speed is immediately stabilized when the robot comes to the flat surface. We guess that this is due to the natural system dynamics of legs (virtual spring and damper leg). The role of VBLA algorithm in swing phase is also important; the robot can choose the appropriate foot position to land with respect to the velocity of CoM.  

It worth mentioning about the measured ground reaction forces in Fig. \ref{fig:GRF}. The robot is free to follow its own dynamics in term of peak ground reaction forces which is similar with the force pattern of human walking or the virtual BTSLIP model. This correlation is a testament to how close the robot is to have the virtual BTSLIP dynamics. 
\subsection{Summary}
In this work, we presented the control strategy for the 5-link robot model using operational space control (OSC) and Polar Jacobian Transpose method. More precisely, swing and stance leg forces are designed based on our previous BTSLIP model control \cite{lee2017biped} and these forces are mapped to joint actuators using OSC  specifically formulated for constrained and underactuated system.  Then, the controller was validated in the dynamical simulation of the model walking on the flat ground and terrain ground, as the model converges to its periodic trajectory from random initial condition within a few steps. Moreover, the performance of the controller was further shown that the walking model was robust against the external force disturbances. The measured ground reaction forces show that developing machines such that they can enforce desired dynamics allow control strategies developed on widely researched fundamental locomotion models to be applied to articulated robots. In next chapter, we will verify how we can apply this control scheme onto the flat foot robot model.
\chapter{Finite State Machine and Hierarchical Control for the 7-link robot model}
\chaptermark{Hierarchical Control Strategy for the Flat Foot model}
\label{chapter:chapter04}
\section{Introduction}
In chapter 3, we have shown how to implement the hierarchical strategy for the point foot model consisting two control layer over uneven terrain and disturbance forces. In this chapter, we extend the point foot model to the flat foot model which weighs 51 kg and has 0.65-m-long legs. The robot is provided no information regarding where the change in height occurs and by how much. 
The hierarchical control strategy will be used for this robot model. By simplify the controller, we consider the virtual BTSLIP model by attaching spring from hip to ankle of each leg and decouple the ankle control with existing controller inherited from Chapter 3.  
In this chapter, the walking cycle is implemented using a Finite State Machine (FSM). By using the local information of leg, the mode control will be conducted by FSM. The 7-link robot model is controlled to have two virtual springy legs which are similar to the design in the BTSLIP model. The stance mode control will be built with appropriate modification from \cite{lee2017biped}. The proportional-derivative (PD) controller and Velocity Based Leg Adjustment are applied to the swing mode control. Additionally, ankle push-off phase of trailing limb during the step-to-step transition in walking will be activated by the decision from FSM. 
\begin{figure}[ht!]
    \centering
    \includegraphics[width=0.8\textwidth,height=0.9\textheight,keepaspectratio]{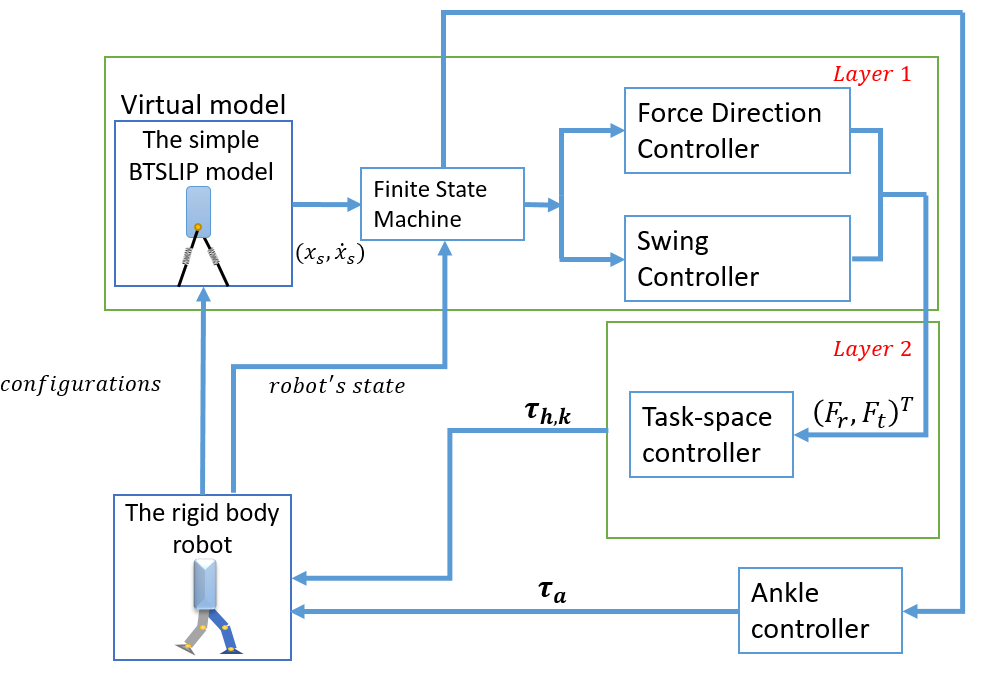}
    \caption{Block diagram of the simple-model-guided task space control system. We combine the hierarchical structure from chapter 3 and the ankle control. Then, the Finite State Machine is designed that manages transitions among controllers. In the simulation, we also test the walking motion under disturbance forces and rough terrain}
    \label{fig:blockdiagram_foot}
\end{figure}
Fig. \ref{fig:blockdiagram_foot} provides an overview of the ``simple-model-guided task space control framework". The details in this figure will be discussed in more detail in the following sections.

The remainder of this chapter is organized as follows. We first present brief descriptions of the robot model and the ODE integrator in Section 4.2. The control strategy for the robot model is highlighted in Section 4.2. Section IV presents the simulation results as the performance of the control strategy. Finally, we conclude the paper with final remarks in this chapter. 
\begin{table}[h]
\normalsize
\caption{Model parameters}
\label{table_example}
\begin{center}
\begin{tabular}{| m{2cm} | m{1cm}| m{1cm} | m{1cm} |m{1cm}|}
\hline    
Model parameters & Trunk (t) & Femurs (f) & Shins (s)& Foot (a) \\
\hline
\(m_{*}\) (\(kg\)) & 27.13 & 3.94 & 2.38 & 2.86 \\ 
\(L_{*}\)(\(m\)) & 0.5 & 0.28 & 0.28 & 0.125\\
\(J_*\) (\(kg m^2\)) & 0.58 & 0.014 & 0.03 & 0.01 \\
\hline
\end{tabular}
\end{center}
\end{table}
\section{Robot model}
  
\begin{figure}[h!]
    \centering
    \includegraphics[width=\textwidth,height=0.9\textheight,keepaspectratio]{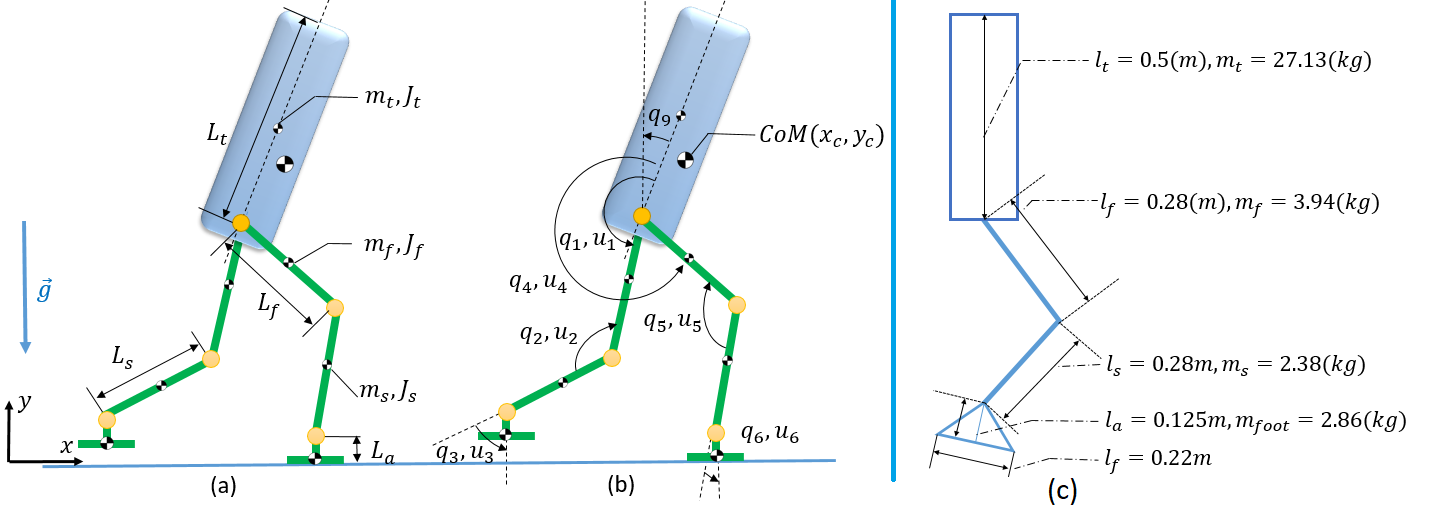}
    \caption{The robot model parameters (a). The robot model coordinates follow the convention such that angles and torques are defined positive in counter-clockwise direction (b). \(q_9\) is the absolute trunk angle. The position of the floating trunk are denoted as \([x_c,y_c]\)). \(\) in Cartesian coordinates.The size of the biped robot in Open Dynamics Engine (c). This parameters of robot's size is adopted from the Mahru robot in 2D version.}
    \label{fig:first}
\end{figure}
The robot model in this paper consists of seven links as depicted in Fig. \ref{fig:first} a trunk and two identical lower limbs with each limb having a femur, a shin, and an ankle; moreover, all links have mass and are connected by hinge joints. The robot parameters are shown in Table. I. All of the joints are considered only rotating in the sagittal plane. With 7-link, the model in sagittal plane consists of 9 Degree of Freedom (DoF). Six actuated DOF associated with the joint coordinates. Two underactuated DOFs related to the horizontal and the vertical displacements of the Center of Mass (CoM). The last one underactuated DOF associated with the orientation of the robot in a sagittal plane. Thus, the generalized coordinates of the system ($q_e$) can be combined by two subsets q and r. 
\begin{equation*}
\label{eqn: coordinates}
q_e := [q,r]^T 
\end{equation*}
where \(q:= [q_1,q_2,q_3,q_4,q_5,q_6.q_9]\) encapsulates the joint coordinates, \(q_9\) is the underactuated DOF which associated with orientation of the robot. Besides, \(r:=[x_c,y_c] \in \mathbb{R}^2\) is the Cartesian coordinates of the CoM of robot. 
\begin{table}[h]
\normalsize
\caption{Open dynamics engine parameters}
\label{table ODE}
\begin{center}
\begin{tabular}{| m{6cm} | m{3cm} |}
\hline    
Name & Value \\
\hline
Time of Step & 0.001$[s]$ \\ 
Gravity & -9.81 $[m/s^2]$\\
CFM & $10^{-5}$ [$\cdot$] \\
ERP & 0.9 [$\cdot$]\\
Contact Surface Layer & 0.001 [$\cdot$]\\
ContactMaxCorrectingVel & 100 [$\cdot$]\\
\hline
\end{tabular}
\end{center}
\end{table}
We use Open Dynamic Engine (ODE) for simulating rigid body dynamics and deactivate the motion in Z plane to force the movement only in the sagittal plane. The ODE uses Newton-Euler approach to describe the system of bodies by a full set of coordinates, and set of constraining equations. In general, the state of each body part in ODE is described
by six variable $x = [p_x \: p_y \: p_z \: \theta_x \: \theta_y \: \theta_z]^T$ containing the position and orientation of the body, and an additional function h(x) that describes the allowed motion for the body. Using Newton's second law, a system of n bodies can be described as:
\begin{equation}
\label{ICMA:eq1}
\textbf{F}_{external} + \textbf{F}_{constraint} = M\mathbf{\ddot{x}},
\end{equation}
where $\mathbf{\ddot{x}} = [\mathbf{\ddot{x}}_1,\mathbf{\ddot{x}}_2,...,\mathbf{\ddot{x}}_n]$ is the acceleration of \(n\) rigid bodies, \(M\) is body mass matrix, and force $\textbf{F}$ is decomposed into external and constraint force. ODE uses the first-order
approximation by the discretization with time step $\Delta t$: $\ddot{x} \approx \dfrac{\dot{x}^{n+1} - \dot{x}^n}{\Delta t}$. By substituting the approximation in \ref{ICMA:eq1} and rearranging yields: 
\begin{equation}
\label{ICMA:eq2}
M^{-1}\textbf{F}_{constraint} = \dfrac{\mathbf{\dot{x}}^{n+1}}{\Delta t} - (\dfrac{\mathbf{\dot{x}}^n}{\Delta t} + M^{-1}\textbf{F}_{external}).
\end{equation}
Corresponding constraint forces are necessary to ensure that
the system follows the constraints. The direction of these
forces is known and perpendicular to the allowed motion, it
can be expressed by a Jacobian
\begin{equation}
\label{ICMA:eq3}
\textbf{F}_{constraint} = J^T \boldsymbol{\lambda},
\end{equation}
where $\lambda$ are the unknown signed magnitudes of the constraint forces, called Lagrange multipliers. In ODE system, the contact constraint are designed to allow some natural penetration of two soft, bouncy object, hard constraint, and to compensate the drift position error between two objects. To allow implementation of those constraints, ODE yields the velocity constraints at the (n+1) step as below:
\begin{equation}
\label{ICMA:eq4}
\mathbf{\dot{x}}^{n+1} = J^{-1}(c-k_{cfm}\boldsymbol{\lambda})
\end{equation}
where $k_{cfm}$ is a square diagonal matrix that mixes the constraints with the corresponding constraint forces, c is the corrective term for error position. To determine the value of constrain force, we substitute the relations for the constraint forces (\ref{ICMA:eq3}) and velocity constraints (\ref{ICMA:eq4}) in (\ref{ICMA:eq2}) and rearrange the formulation as below:
\begin{equation}
\label{ICMA:eq5}
(JM^{-1}J^T + \dfrac{1}{\Delta t} K_{cfm})\boldsymbol{\lambda} = \dfrac{c}{\Delta t} - J(\dfrac{\mathbf{\dot{x}}^n}{\Delta t} + M^{-1}\textbf{F}_{external}).
\end{equation}
Furthermore, ODE has boundary on the constraint force:
$\lambda \geq 0$ and friction coin rule to prevent movement in the
tangential direction: $\lambda \leq \mu F_n$ with $F_n$ is the normal
force. With boundary condition above and (\ref{ICMA:eq5}), the constraint force can be solved by the Linear Complementary Problem (LCP), 
\begin{equation}
\label{ICMA:eq6}
\begin{aligned}
& A\boldsymbol{\lambda} = b \\ 
& \boldsymbol{\lambda}_{min} \leq \boldsymbol{\lambda} \leq \boldsymbol{\lambda}_{max}, 
\end{aligned}
\end{equation}
where $A = (JM^{-1}J^T + \dfrac{1}{\Delta t}K_{cfm})$ and $B = \dfrac{c}{\Delta t} - J(\dfrac{\mathbf{\dot{x}}^n}{\Delta t} + M^{-1}\textbf{F}_{external})$. ODE uses the projected Gauss-Seidel (PGS) with Successive Over-Relaxation (SOR). Open dynamic engine parameters are given in Table. II.
\begin{table}[h]
\caption{Control parameters for flat foot robot model}
\label{table:control}
\begin{center}
\begin{tabular}{|c|c|c|c|c|}
\hline
Parameter & Meaning  & Value [unit]\\
\hline
$k$ & spring stiffness for virtual leg & 15000 [$N/m$]\\
$k_d$ & virtual spring damping & 150 [$N\cdot s /m$]\\
$k_a$ & ankle control position-proportional gain & 15 [$N/m$]\\
$c$ & position-proportional gain & 10 [$\cdot$] \\
$d$ & velocity-proportional gain  & 1 [$s$]\\
$\mu$ & coefficient value for VBLA & 0.6 [$\cdot$] \\
$L_0$ & the rest length of virtual leg & 0.55 [$m$]\\ 
$\Delta L_d$ & the maximum of retracted length & $0.2L_0[m]$\\
$\Delta \psi$ &the deviation of $\psi$ & $\pi/3[rad]$\\
$\tilde{\psi}_d$ & the desired trunk angle & $-\pi/40[rad]$ \\
$q_a^{sw}$ & the desired ankle angle in swing phase & $\pi/6[rad]$ \\
$q_a^p$ & the desired ankle angle in push-off & $-\pi/9[rad]$\\
\hline
\end{tabular}
\end{center}
\end{table}
\section{Control Strategy} 
In this section, the control strategy for robot model is described. First of all, we introduce the state machine for walking motion. In this way, the walking state machine will determine controller mode for each leg at each walking phase. To stable the trunk, we embed the simple force direction control in our previous paper \cite{lee2017biped} with appropriate modification to the robot model. In Fig. \ref{fig:control}, the control variables associated with stance and swing phase are shown. Herein, we create a virtual spring between hip joint and ankle joint of two legs, as utilized in the BTSLIP model. The hip torque and knee torque are mapped as close as possible to the simple model while the ankle torque follows the command from walking state machine to active push-off phases and heel-strike phases. 
\subsection{A Finite State Machine}
\begin{figure*}[h!]
    \centering
    \includegraphics[width=\textwidth,height=12cm,keepaspectratio]{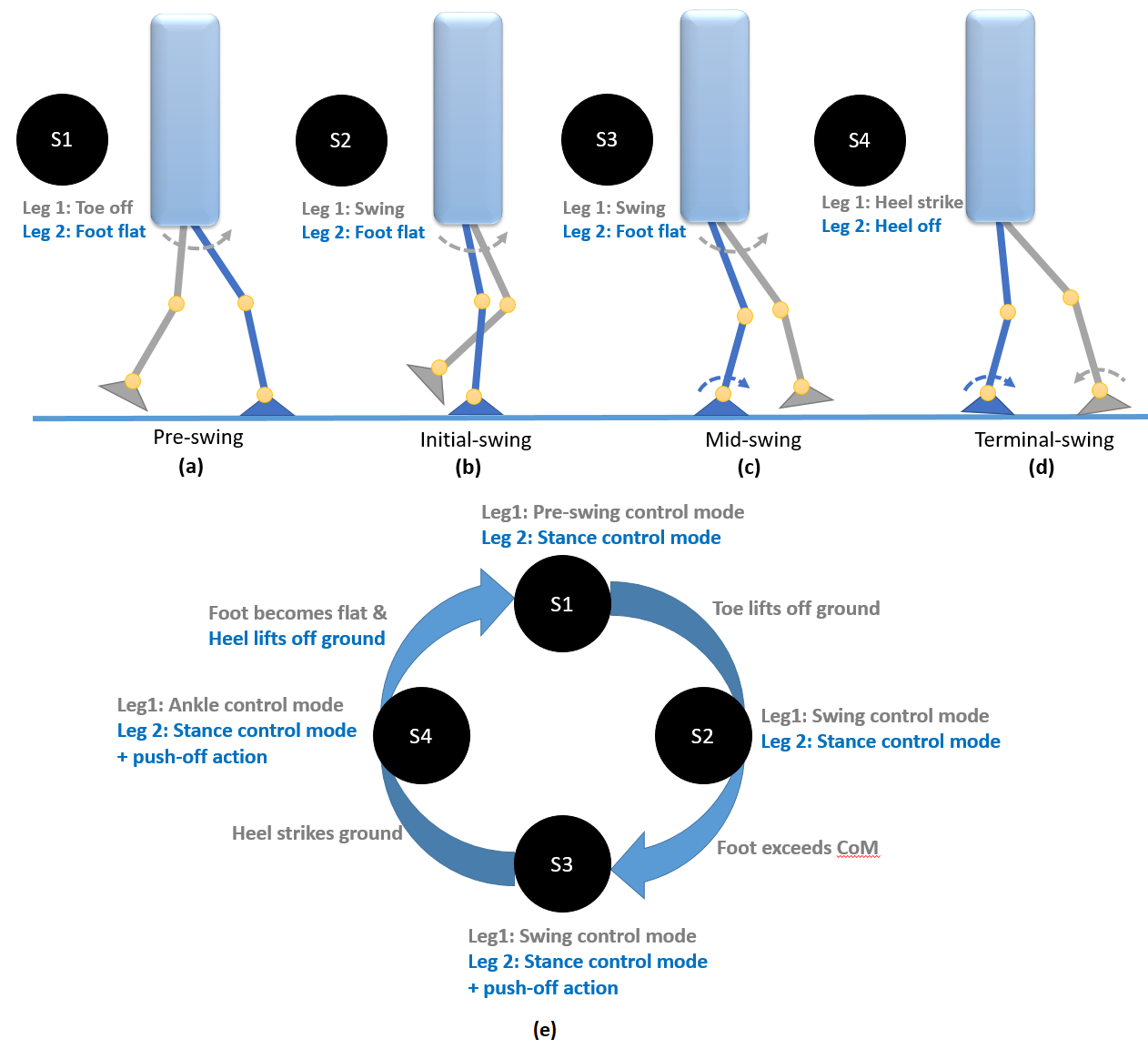}
    \caption{The gait phases upon which the state machine is based. The labels correspond to the phase of the gray leg (leg 1) (a) - (d). A Finite State Machine is used for each leg to determine controller mode for walking phases (e). The leg 1 and 2 are marked with gray color and blue color. S1 (a) is the pre-swing phase. The stance mode control is applied to the leg 2, and the swing mode control is applied to the opposite leg. S2 (b) is the initial swing phase. The leg 1 and 2 are still in swing mode and stance mode control. Ankle push-off control will be applied to leg 2 if $p_{toe} \leq p_{CoM}$, where $p_{toe}$ and $p_{CoM}$ are the toe horizontal position of the leg 2 and horizontal position of CoM. S3 (c) is the mid-swing phase. The leg 2 is still in the stance mode control and push-off mode control. S4 (d) is the terminal-swing phase, heel of the leg 1 touches the ground while the heel of the leg 2 takes off from the ground. In this sate, the leg 2 quickly turns to swing mode control.} 
\label{ICMA:FSM}
\end{figure*}

It is worth to note that each leg has specific mode control in each particular phase that defines in Finite State Machine, and does not require the information of the opposite leg. Therefore, without loss of generality in Figure. \ref{ICMA:FSM}, we start the description for a waking state machine with a toe-off event at the leg 1. According to Figure. 2:

$\bullet$ \textit{State 1 (S1):} The controller for the leg 1 is in swing mode control while the controller for the leg 2 is in the stance mode control. Ankle angle of the swing leg will be controlled to a fixed angle until it touches the ground by its heel in the last state. 

$\bullet$ \textit{State 2 (S2):} The swing leg (i.e. leg index 1) will be retracted and swung forward by swing mode control. The stance mode control is still applied to the leg 2. When the horizontal position of Center of Mass exceeds the toe of the leg 2, the ankle control will be applied to activate ankle push-off \cite{lipfert2014impulsive} to the existing stance leg. 

$\bullet$ \textit{State 3 (S3):} The leg 1 is still in swing mode control. The ankle push-off control and stance mode control are applied to the leg 2. 

$\bullet$ \textit{State 4 (S4):} A touchdown event for the swing leg (i.e. the heel of leg 1 touches the ground) will lead the system to this state. The controller will apply the ankle torque to the leg 1 to dampen the impulsive impact until its foot becomes flat. Instantaneously, FSM starts again with the correspondence swing leg 2. 

Additionally, the condition of foot will be determined as follow: (1) heel-strike (i.e. $p_{heel} = 0$ and $p_{toe} > 0$), (2) foot-flat (i.e. $p_{heel} = 0$ and $p_{toe} = 0$), (3) heel-off (i.e. $p_{heel} > 0$ and $p_{toe} = 0$), (4) toe-off (i.e. $p_{heel} > 0$ and $p_{toe} > 0$), where $p_{heel}$ and $p_{toe}$ are the position of heel and toe in Cartesian coordinate
\subsection{stance mode control} 
\begin{figure}[ht]
    \centering
    \includegraphics[width=0.8\textwidth,height=0.85\textheight,keepaspectratio]{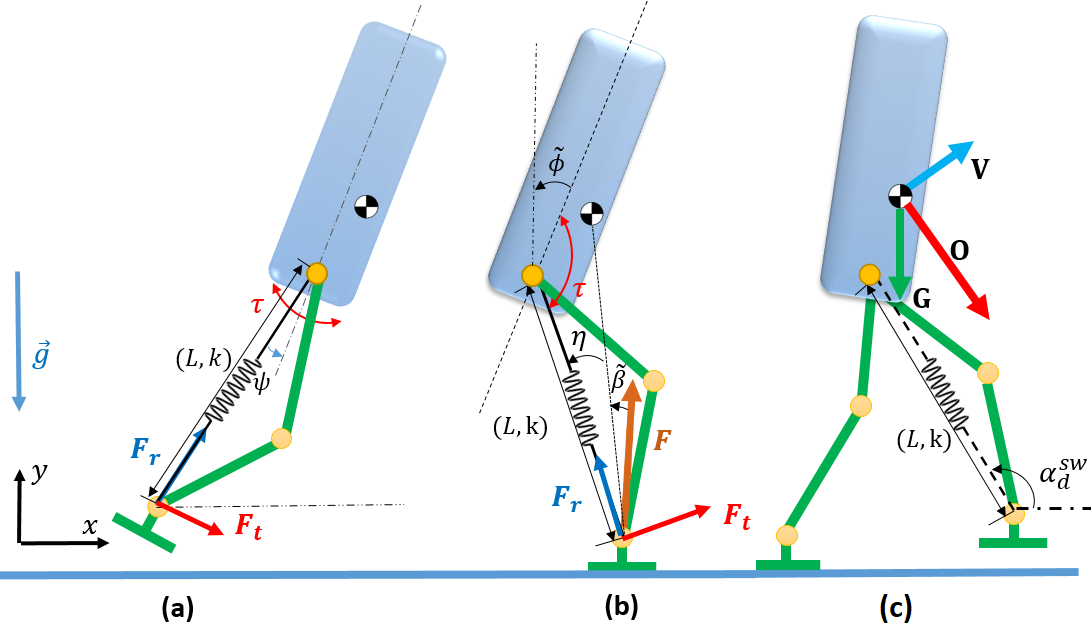}
    \caption{The robot model is considered to have virtual compliant legs. Two virtual spring is connected from hip to each foot with the same spring stiffness, \(k\). The control variables for swing leg (a). And the control variables for stance leg (b). The Velocity based Leg Adjustment \cite{sharbafi2016vbla} (c)}    
\label{fig:control}
\end{figure}
Fig. \ref{fig:control} shows that the 7-link model will have virtual springs from hip to each foot. We inherit the controller of the BTSLIP model [17] in order to calculate the virtual axial force and tangential force that is necessary to maintain the upright trunk in walking for the stance leg. The commanded ankle-end is 
calculated as:
\begin{equation}
\label{ICMA:control}
\begin{aligned}
\textbf{F} = 
\begin{bmatrix}
F_r \\
F_t
\end{bmatrix} 
= 
\begin{bmatrix}
k (L_0 - L) + k_d\dot{L}\\
F_r \tan \beta
\end{bmatrix}
\end{aligned}
\end{equation}
where \(k\) and \(L_0\) are the spring stiffness and the rest length of virtual leg, respectively.  \(L = | p_a - p_h|\) is current leg length (i.e. Virtual leg length is calculated by the distance from ankle to hip), \(p_a\) and \(p_h\) are the position of ankle and hip in global frame, respectively. $\beta = \eta + \tilde{\beta}$ is the angle between \(\textbf{F}\) and the virtual leg, and $\eta$ is the angle between the vector from stance ankle to CoM and virtual leg. We added the damping component in computing of the axial force. Note that we set up the lower limit of \(F_r\) to zero. 

The ground reaction force vector \(\textbf{F}\) is the combination of the virtual spring force and reaction force : $\textbf{F} = \textbf{F}_r + \textbf{F}_t$. The direction of $\textbf{F}$ is controlled by $\tilde{\beta}$. We take into account the speed and direction of the trunk in determining the direction of \(\textbf{F}\) as follows: 
\begin{equation}
\begin{aligned}
&\tilde{\beta} = -c\tilde{\phi} - d\dot{\tilde{\phi}}
\end{aligned}
\end{equation}
where $c$ and $d$ are the control parameters, as shown in Table \ref{table:control}. $\tilde{\phi}$ is the angle of the trunk w.r.t to vertical line.

\subsection{swing mode control} 
We use the method velocity-based leg adjustment (VBLA \cite{sharbafi2016vbla}) for swing leg control, to define the touchdown angle for swing leg, as shown in Fig. \ref{fig:control}(c). The VBLA determines desired swing leg orientation as follows, 
\begin{equation}
\textbf{O} = \mu \textbf{V} + (1-\mu)\textbf{G},
\end{equation}
where \(\textbf{V} = \dfrac{\textbf{v}}{\sqrt{gL}}\) and \(\textbf{G} = \dfrac{\textbf{g}}{|g|}\) are non-dimensionalized CoM velocity and gravitational acceleration, respectively. $\textbf{v}$ and $\textbf{g}$ are the vector of CoM velocity and the gravity. The angle between the desired vector \(\textbf{O}\) and the ground is determined as the desired touch down angle \(\alpha_d\). By defining the virtual leg length trajectory for swing leg \(L_d\), and combining with the desired touch down angle from VBLA, the simple form of proportional - derivative control can be applied for swing leg control: 
\begin{equation}
\begin{aligned}
\textbf{F}
\begin{bmatrix}
F_r \\
F_t
\end{bmatrix}
=
\begin{bmatrix}
k(L_d - L^{sw}) + \dfrac{k}{10^2}(\dot{L}_d - \dot{L}) \\
(c(\alpha_d - \alpha) + \dot{\psi})/L
\end{bmatrix}
\end{aligned}
\end{equation}
where $\alpha$ is the angle created by the virtual swing leg and the horizontal axis. The virtual leg length can be retracted to prevent the scuffing of the leg with the walking surface by
controlling the desired swing leg length $L_d$:
\begin{equation} 
\begin{aligned}
\left \{
\begin{array}{ll}
 L_d = 0.8L_0 + \dfrac{\Delta L_d}{2} - \dfrac{\Delta L_d}{2} \cos(\dfrac{\tilde{\psi}2\pi}{\Delta\psi}) \\ \: \textbf{if} \: \tilde{\psi} \geq \dfrac{\Delta\psi}{2} \:\textbf{or}\:\textbf{if} \:\tilde{\psi} \leq \dfrac{-\Delta\psi}{2}\\
 \textbf{then} \:L_d = L_0, 
 \end{array}
\right.
\end{aligned}
\end{equation}
where $\tilde{\psi} = \psi + \tilde{\psi}_d$. $L_d$ and $\Delta \psi$  are control parameters,
as shown in Table III. The stance and swing control set the commanded forces at the ankle in axial and tangential direction. This set of forces can be turned into hip and knee torques $\textbf{u} = [u_h,u_k]^T$ via Jacobian $\textbf{J}_{local} \in \mathbb{R}^{2 \times 1}$  from hip to ankle for each leg.
\begin{equation}
\textbf{u} = \textbf{J}_{local} \textbf{F}
\end{equation}
\subsection{Ankle mode control}
Feet and ankles support many benefits to bipedal walking. It can help to reduce the fluctuations of velocity when the center of pressure on the foot can travel forward. They also help to reduce the impulsive impact then the heel strikes the ground and to inject energy at the end of the stride through toe off. According to FSM, the torque at the ankle can be controlled actively. In order to dampen impulsive velocity when heel strikes the ground, the ankle torque can be described as below: 
\begin{equation}
u_a = -\dfrac{k_a}{10}\dot{q}_i
\end{equation}
where index $i = \{3,6\}$ depends on touching ground condition of legs. When the leg satisfies the condition of push-off in the FSM, the ankle control tries to plantar flex the ankle angle. The simple PD control for the ankle in this phase is expressed as below: 
\begin{equation}
u_a = k_a q_{ankle}^p - \dfrac{k_a}{10}\dot{q}_i,
\end{equation}
and during toe off state (i.e. Swing phase), the ankle is served to a fixed angle using a PD controller: 
\begin{equation}
u_a = k_a(q_i - q_{ankle}^{sw}) - \dfrac{k_a}{10}\dot{q}_i, 
\end{equation}
where $q_{ankle}^p$ and $q_{ankle}^{sw}$ ankle are control parameter as shown in Table. III.
\begin{figure*}[h]
    \centering
    \includegraphics[width=\textwidth,height=7cm]{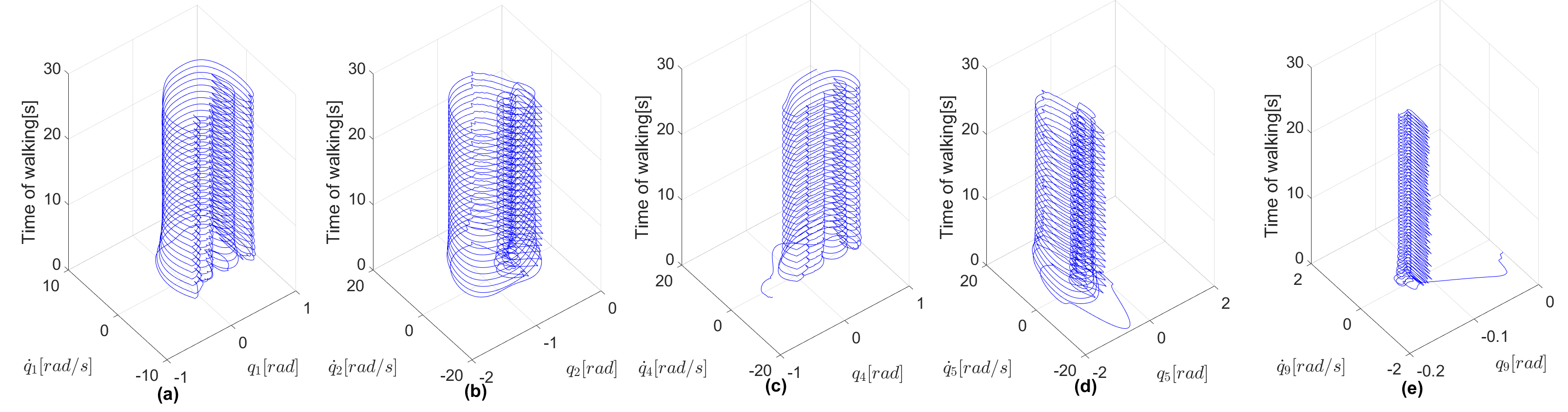}
    \caption{The model begins with a random initial condition and converges steady motion. The phase portrait of an angle from trunk to the thigh of the leg 1 (a) and leg 2 (b). The phase plot of stance knee angle of the leg 1 (c) and the knee angle of the leg 2 (d). The
phase plot of absolute trunk angle (e).} 
\label{fig:result1}
\end{figure*}
\section{Simulation Result and Discussion}
\begin{figure}[ht]
    \centering
    \includegraphics[width=\textwidth,height=0.4\textheight]{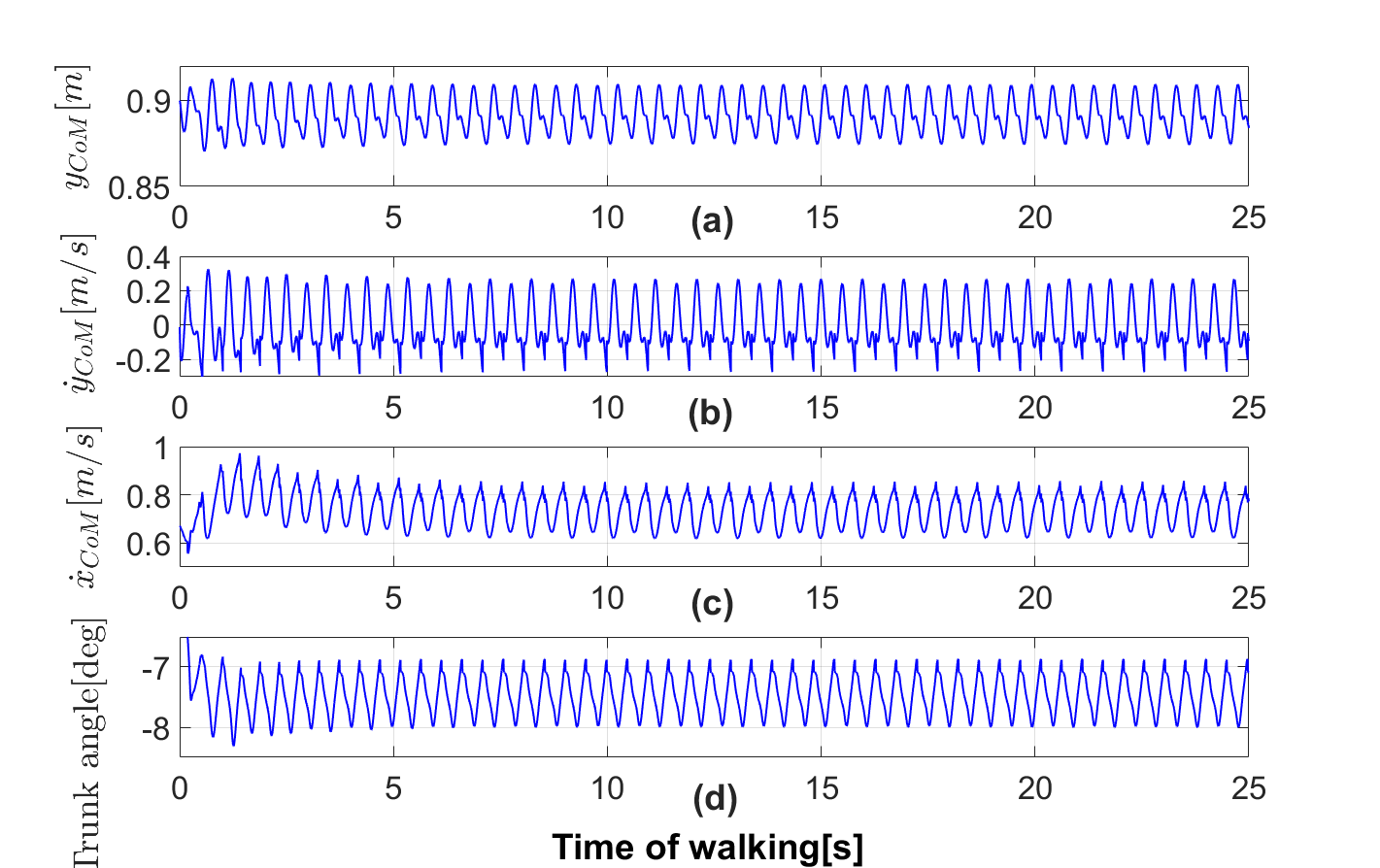}
    \caption{ Trajectories in 25 seconds of the 7-link robot model. The vertical position of the CoM (a). The velocity of vertical position of CoM (b). The forward velocity of the robot model (c). The pitch of trunk is plotted in (d)}
\label{fig:general}
\end{figure}
\begin{figure}[h!]
    \centering    \includegraphics[width=\textwidth,height=0.4\textheight,keepaspectratio]{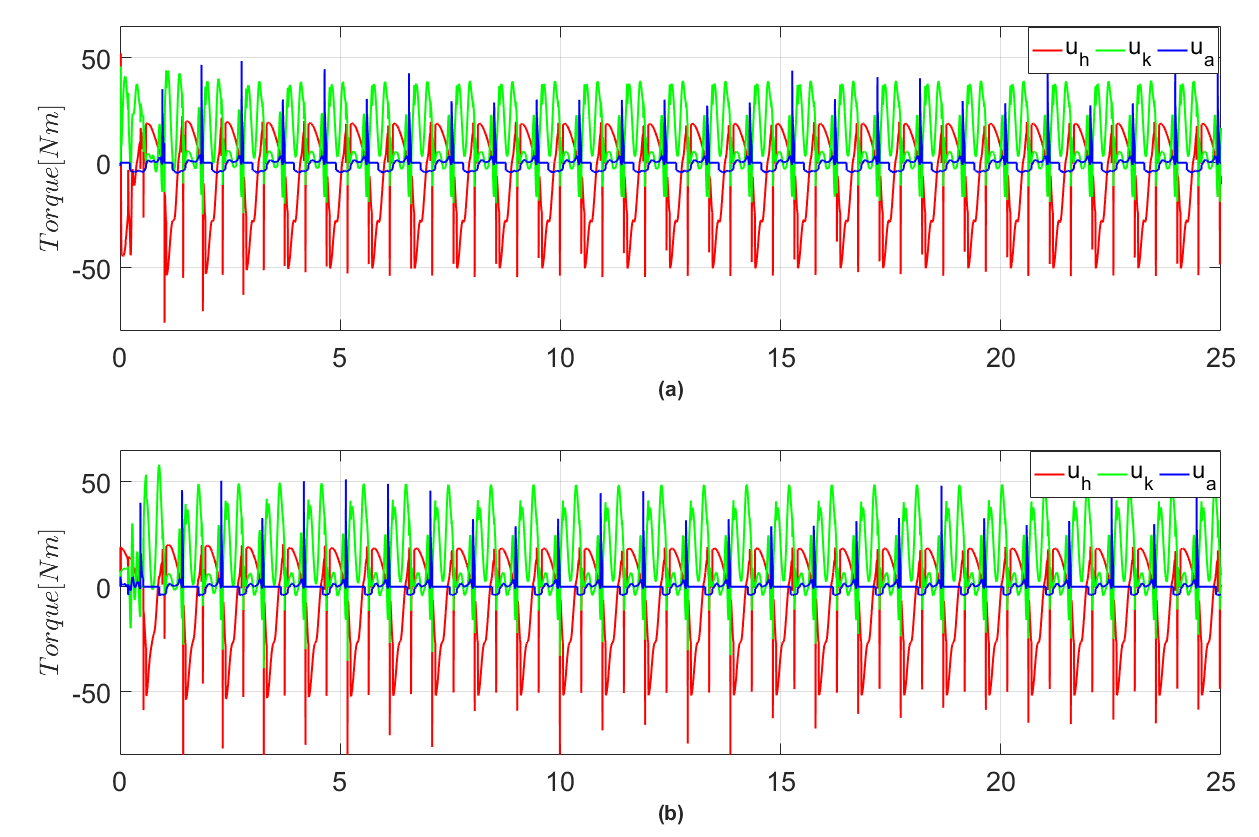}
    \caption{ The control inputs of walking motion in 30 seconds. The hip torque ($u_h$ : blue), knee torque ($u_k$ : green), and ankle torque ($u_a$: blue) are applied to leg 1 (a), and leg 2
(b).}   
\label{res:control}
\end{figure}
\begin{figure*}[h!]
    \centering
    \includegraphics[width=11cm,height=8cm]{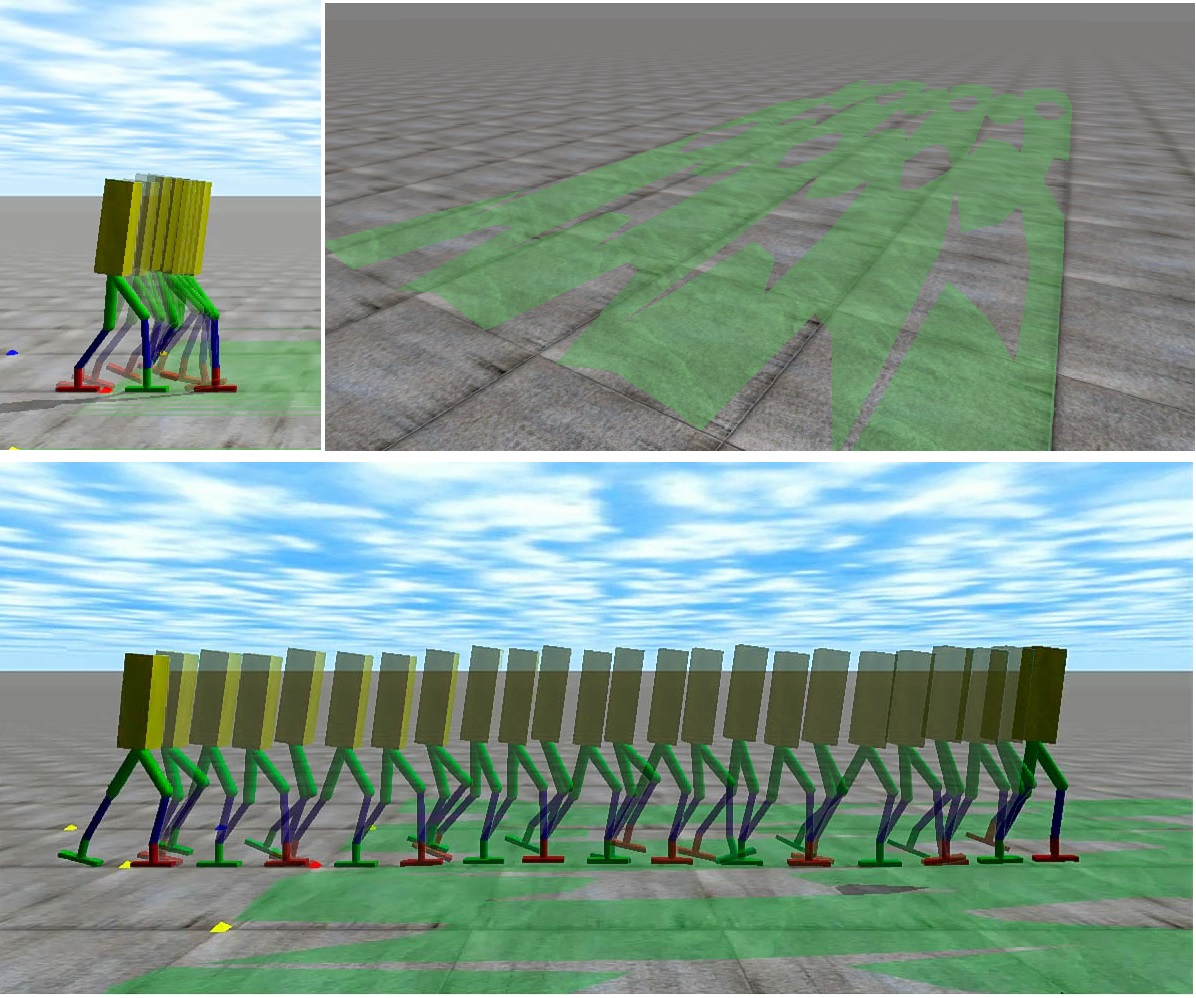}
    \caption{The snapshots on the above left are captured in 0.4 seconds and show one swing phase motion. The rough surface is shown in the above right side with the range of height about $[0:5]cm$. The snapshots below show several steps in approximately 9 seconds on terrain ground.} 
\label{fig:natural walking}
\end{figure*}
To test the validity of the proposed control strategy with the 7-link rigid model, we use the dynamic simulation in ODE environment. The model parameters, environment parameters, and control parameters are listed in Table I, Table II and Table III, respectively. The 7-link robot dynamics are illustrated in Section II, and the proposed control is presented in Section III.

The trajectories of $y_{CoM}$, $\dot{y}_{CoM}$, $x_{CoM}$, and $q_9$ are plotted in Fig. \ref{fig:general}. The beginning of trajectories due to the initial conditions has jerky motions, and it changes to stable motion around 1:5[s]. A simulated robot walked at a moderate speed (approximately $0.75 [m/s]$). It is worth noting that the control algorithm does not use any explicit speed control method, but the speed is stabilized. We intuitively speculate that this is because of the natural system dynamics, in the same way, that speed is naturally stabilized by the VBLA controller and springy legs mechanism. Additionally, in Fig. 6(c), the trunk angle of the robot model oscillates around the desired angle $\tilde{\phi}_d$ 

It can be seen in Fig. \ref{res:control}  that the control signals oscillate with large range at the very first steps. After that, the steady state behavior of control inputs is attained with the range of hip torques are about \([-50,20]^T\)[Nm] and \([-20,40]^T\)[Nm] for the knee torques. In the ankle push-off phase, the instantaneous ankle torques will be applied which results as spikes in Fig. \ref{res:control}.  
\begin{figure}[h!]
    \centering
    \includegraphics[width=\textwidth,height=0.4\textheight]{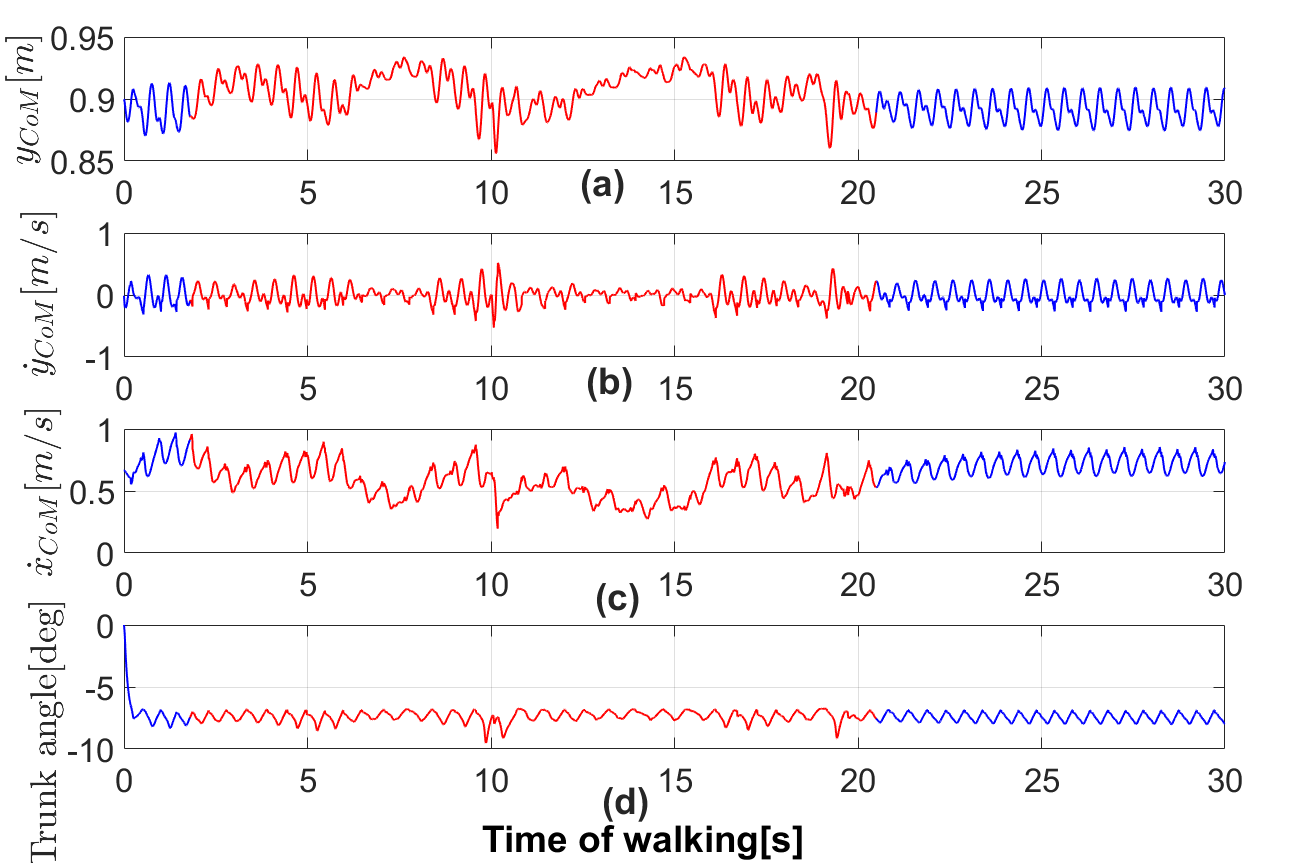}
    \caption{ Trajectories in 30 seconds of the 7-link robot model on rough terrain map with the same initial conditions in Fig. \ref{fig:general}. Vertical position and velocity vertical position of CoM are drew in (a), and (b). The speed of robot and trunk orientation of the robot are in (c) and (d).  The plots are changed color to red when robot is in the terrain ground.}
\label{fig:roughmap}
\end{figure}

To further investigate the performance of control strategy, we simulate the biped robot on the rough terrains ground. We use height map terrains function in ODE, the height level of each point in ground terrain map will be defined by formulation: $z(cm) = 1 + 4\sin(20y+20x)$, where $x$, $y$ and $z$ are the location of each point on the terrain map. More clearly, the height of each point in terrain map will be in the range $[0:5]cm$. The terrain map is illustrated on the right side of Fig. \ref{fig:natural walking}. Additionally, the drawing on the left side of Fig. \ref{fig:natural walking} show completely swing phase of one leg. The robot still keeps a natural walking gait even in the rough terrain surface, as shown in the figure below in Fig. \ref{fig:natural walking}. Fig. \ref{fig:roughmap} shows the robustness of our proposed control against terrain map. The trunk is quickly turned to steady motion when robot overcomes the terrain surface; the jerky behavior in the horizontal and vertical motion of CoM is also diminished after touching to the flat surface. 

\section{Summary}
In this work, we presented the control strategy for the 7-link robot model including the force direction control and simple ankle control. More precisely, the force direction control for hip and knee torques are designed based on our previous control for the BTSLIP model \cite{lee2017biped} and these forces are mapped to joint actuators by Jacobian transpose.
Besides, the swing controller is capable of retracting the swing leg to prevent it being scuffed with the ground. In additionally, the FSM will activate the ankle push-off mode
control in appropriate phases and helps to reduce the impulsive impact by controlling the ankle torque in heel strike phases. The controller was validated in Open Dynamic Engine environment on the flat ground and terrain ground, as the model converges to its steady motion from random initial condition within a few steps. While the results of this work are limited to specific
choices of the model (mass and length, size), we believe that this controller is sufficiently practical to be a basis for future research. Besides, the control for ankle model seems
very simple and needs more research on its. As for our future work, we will focus on extending the articulated robot with more real size and improve the control strategy. Reaching the
walking controller for the frontal balance and turning in the 3-D environment are also of our desire. 

\chapter{Summary and Future Work}
\chaptermark{Summary and Future Work}
\label{chapter:chapter06}
\section{Summary}
This work presents an attempt to control the 2D dynamic walking of a humanoid robot over disturbance forces and rough surfaces based on the reducing ordered model (i.e., the Biped Trunk - SLIP model). The overall control approach is divided into two levels: (1) The high-level planner focuses on the performance of the most emphasized template model for a dynamic walking motion such as the force profile. By studying the stability in terms of robustness to external disturbance forces and the rough surface of the BTSLIP model integrated with the force direction controller, the main dynamic characteristics of this combination are captured. (2) The lower-lever controller, we study two methods in the field of task space controller, the simple Jacobian transpose formulation and the operational space controller. It is worth noting that the simple control method does not require a detailed whole-body model, but relies on the force control capabilities of the robot. The performance is not significantly different from that of operational space control, while the computational cost and complexity of the overall control strategy are significantly reduced. 

In the first chapter, we explore the possible control rules that can be applied to the BTSLIP model. We draw inspiration from the concept of a Virtual Pendulum, which is proposed as an intuitive posture stabilization strategy and successfully demonstrated on the compliant leg scheme. In the first chapter, we first propose a combined control strategy consisting of a discrete linear quadratic controller to adjust the position of the virtual pivot point (VPP) and a linearization controller to change the leg stiffness. In addition, quadratic programming is used to find the periodic solution for the walking model with the VP concept. As a result, the BTSLIP model is well equipped with the combined control strategy. Later, the force direction control is proposed for the BTSLIP model because we want to reduce the computational cost burden of the first method. 
Following the concept of the virtual pendulum, we argue that the direction of the ground reaction force is more important than the location of a VPP point. Interestingly, with this simple control, the walking model can achieve a robust walking motion with no falls. 

In the next chapters, we apply this simple control to the articulated body robot (e.g., the point foot model and the flat foot model). We make some changes to the force direction control and add the state-dependent control for the swing leg. Together with the force direction control to stabilize the walking motion, the state-based control for the swing leg can help the robot avoid rubbing on the ground and adapt to rough terrain based on the speed-based leg adaptation method. Then, the desired force profile is passed to the task space controller to calculate the joint moments of the walking robot. The final result shows that the point-foot model and the flat-foot model run robustly over rough surfaces and disturbance forces. Moreover, the resulting GRFs of the point-foot model are rich in human-like features, such as the double-tip GRF pattern. 
\section{Future work}
The development in this work has brought some clues for many future research directions. First, the change in swing leg control strategy can adjust and track the speed of bipedal walking, but it does not perfectly change the speed as close as the set point. We believe that both the adjustment of foot position in swing leg control mode and the control of leg length in stance leg control mode help to accomplish the task. On the other hand, we would improve the current balance control by researching some existing methods such as Capture Point, and Divergent Component Points in 3D. Although this is likely to be a motion planning-based control, we would find a way to improve our state-based control. Finally, we are also very interested in extending the work to 3D robot models. 

\appendix        													%%  This command is needed before you start inserting your 
																	%%	appendices for proper formatting.
%\insertChapter{"Appendix_01"}

% These two last commands are not to be changed either. The first one prepares the bibliography and sets its style while the second inserts the author's vita.
\insertBibliography
%\insertVita{vitaFile}
\end{document}